\documentclass{article}

     \PassOptionsToPackage{numbers, compress}{natbib}

\usepackage[preprint]{neurips_2026}

\usepackage[utf8]{inputenc} 
\usepackage[T1]{fontenc}    
\usepackage{hyperref}       
\usepackage{url}            
\usepackage{booktabs}       
\usepackage{amsfonts}       
\usepackage{amsmath}        
\usepackage{amssymb}        
\usepackage{nicefrac}       
\usepackage{microtype}      
\usepackage{xcolor}         
\usepackage{graphicx}       
\usepackage{algorithm}      
\usepackage{algorithmic}    
\usepackage{multirow}       
\usepackage{subcaption}     
\usepackage{placeins}

\title{EMAG: Differentiable 4D Gaussian Mixture Splatting\\for EEG Spatial Super-Resolution}

\author{
    Alex Lazarovich \quad Ofir Itzhak Shahar \quad Gur Elkin \quad Ohad Ben-Shahar \\
    Stein Faculty of Computer and Information Science \\
    Ben-Gurion University of the Negev, Israel \\
    {\tt\small \{alexlaz,   shofir, gurshal\}@post.bgu.ac.il, ben-shahar@cs.bgu.ac.il}
}

\begin{document}

\maketitle

\begin{abstract}
High-density electroencephalography (HD-EEG) enables fine-grained measurement of cortical activity but requires expensive hardware and lengthy setup times, limiting its clinical and research accessibility.
We propose \textbf{EMAG} (\textbf{E}EG \textbf{M}ixture of \textbf{A}nisotropic \textbf{G}aussians), a differentiable framework that reconstructs HD-EEG signals from a sparse subset of low-density (LD) electrodes by representing brain electrical sources as a mixture of anisotropic 4D space-time Gaussians.
EMAG places a mixture of multiple Gaussians at each point of a spherical brain grid, each parameterized by a full $4 \times 4$ precision matrix, enabling anisotropic spatial spreads and explicit coupling between spatial and temporal dimensions.
The forward model renders scalp EEG via differentiable Gaussian field contributions at electrode locations, enabling end-to-end training without explicit source localization supervision.
We evaluate EMAG on three public EEG benchmarks (Localize-MI, SEED, and SEED-IV) at super-resolution factors of $2\times$ through $8/16\times$.
EMAG outperforms the current state-of-the-art EEG super-resolution method at most super-resolution factors on three standard benchmarks (Localize-MI, SEED, SEED-IV).
The explicit Gaussian parameterization further enables direct visualization and interpretability of learned brain source configurations, potentially opening avenues for clinical and neuroscientific applications, such as source localization or biomarker discovery.
\end{abstract}

\section{Introduction}
\label{sec:intro}
\begin{figure*}[t]
  \centering
  \caption{Overview of EMAG. (1) LD electrodes drive a 1D TemporalEncoder + MLP that outputs amplitude adjustments. (2) The brain volume hosts $G$ anisotropic 4D Gaussians per spherical-grid point. (3)~Differentiable rendering at electrode positions yields the HD scalp signal $\hat{\mathbf{X}}^{\mathrm{HD}}$.}
  \includegraphics[width=0.9\textwidth]{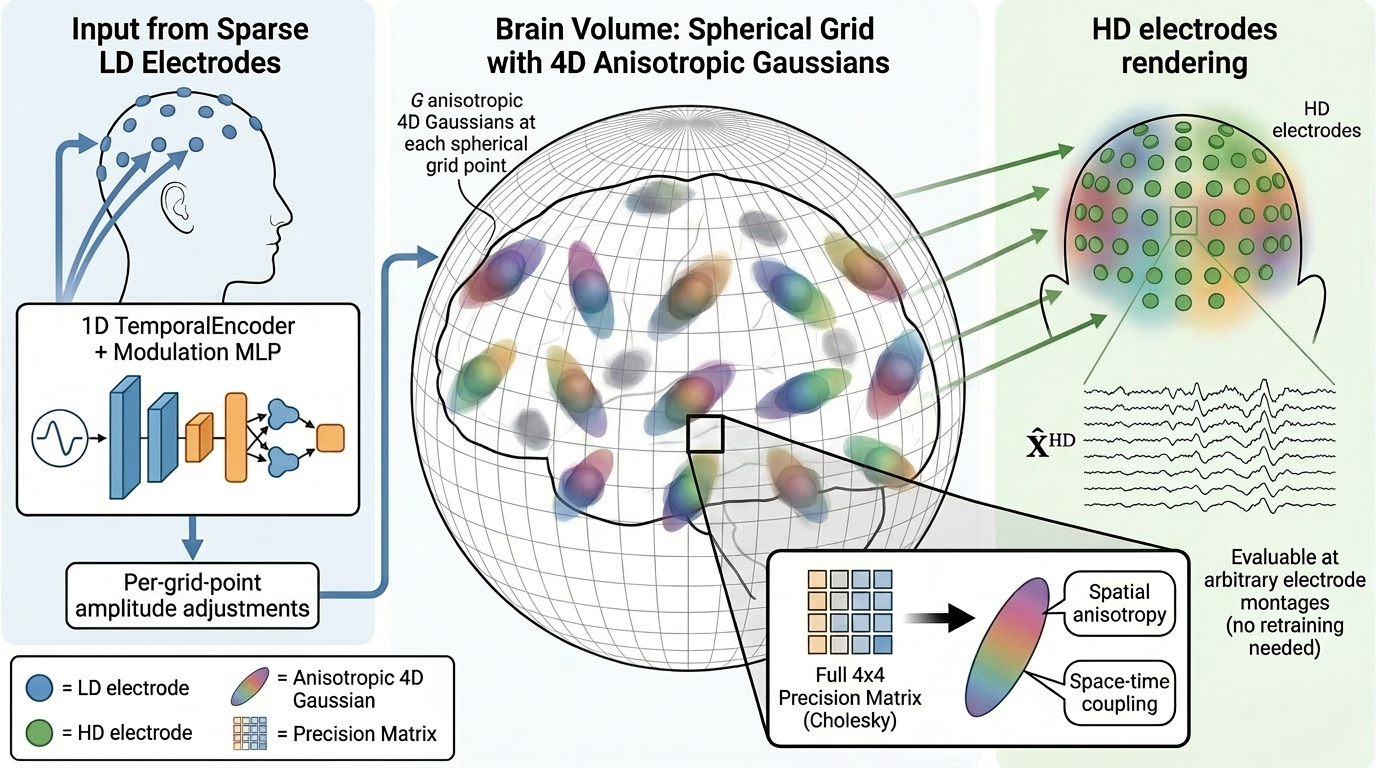}
  \label{fig:concept}
\end{figure*}

Electroencephalography (EEG) is the most widely used non-invasive neuroimaging modality for measuring cortical electrical activity with millisecond temporal resolution~\citep{ramadan2017brain, wolpaw2007brain}.
While standard clinical EEG employs 19--32 electrodes, high-density (HD) systems with 128--256 channels are increasingly recognized as essential for accurate source localization~\citep{grech2008review} and brain-computer interfaces~\citep{lawhern2018eegnet}.
And yet, such HD-EEG acquisition remains impractical in many settings due to the cost of electrode caps, extended preparation time, and patient discomfort.

Given this gap, {\em EEG super-resolution}, the problem of reconstructing HD-EEG from a sparse low-density (LD) subset, has emerged as a practical alternative~\citep{corley2018deep}.
Existing approaches treat EEG super-resolution as a channel-to-channel mapping problem, using convolutional, adversarial, or diffusion-based architectures that operate directly in electrode space\citep{corley2018deep,liu2025step,wang2025generative}.
None of these methods incorporates an explicit forward model of how brain sources project onto scalp electrodes, forgoing the physical inductive bias that could improve both reconstruction accuracy and interpretability.

Unrelated to EEG analysis, the computer vision community has developed powerful differentiable rendering frameworks, most notably Neural Radiance Fields (NeRF)~\citep{mildenhall2021nerf} and 3D Gaussian Splatting~\citep{kerbl20233d}, that represent volumetric scenes as continuous fields and render them via differentiable forward models.
Recent extensions to 4D Gaussian Splatting~\citep{duan20244d, wu20244d} further incorporate temporal dynamics via deformation fields or native 4D covariance matrices.

Recent work on Neural Brain Fields (NBF)~\citep{kedem2025neural} has drawn on the NeRF analogy for EEG, using implicit neural representations to generate signals at unseen electrode positions.
However, NBF uses a per-sample Multilayer Perceptron (MLP) to encode the electromagnetic potential as a continuous function of 3D position and time, without explicit source modeling, and remains fundamentally non-interpretable in terms of neural sources, as it offers no mechanism to localize or characterize the underlying brain sources.

While these two fields have been progressing independently, we observe a structural analogy between volumetric scene rendering and the EEG forward problem: both involve a latent volumetric field (radiance vs.\ source current density) that is projected onto a set of sensors (camera pixels vs.\ scalp electrodes).
This motivates our approach, but requires adaptation: while 4D Gaussian Splatting~\citep{wu20244d} handles dynamic scenes via space-time covariance, it lacks EEG's physics-informed forward model (leadfield-like projection). We thus extend it with anisotropic 4D mixtures on a spherical grid and differentiable Gaussian rendering to scalp potentials, enabling end-to-end SR without explicit inverse supervision---outperforming baselines on most settings.

\textbf{Contributions.} We introduce \textbf{EMAG} (\textbf{E}EG \textbf{M}ixture of \textbf{A}nisotropic \textbf{G}aussians), a framework for EEG super-resolution that advances the application of Gaussian splatting to the neuroscience domain (Fig.~\ref{fig:concept}):
\begin{enumerate}
  \item We formulate EEG super-resolution as \textbf{differentiable rendering of a 4D source field}, where the ``scene'' is modeled as a \textit{mixture} of anisotropic Gaussians placed on a spherical brain grid and the ``rendering'' is a weighted summation of gaussians at electrode positions.
  \item We propose an \textbf{anisotropic 4D Gaussian parameterization} in which each component is equipped with a full space-time $4 \times 4$ precision matrix parameterized via Cholesky decomposition, enabling anisotropic spatial spreads and explicit space-time coupling.
  \item We obtain \textbf{interpretable source representations}: each learned Gaussian carries an explicit 3D position, anisotropic shape, and temporal profile, so post-hoc analysis of the trained model can recover source maps without any source-localization supervision (as shown with Localize-MI at Sec.~\ref{sec:discussion:explainability}).
  \item We provide a \textbf{systematic evaluation} on three public EEG benchmarks (Localize-MI~\citep{mikulan2020simultaneous}, SEED~\citep{duan2013differential, zheng2015investigating}, and SEED-IV\citep{zheng2018emotionmeter}) across super-resolution factors of $2\times$ through $16\times$, including comparison against seven established EEG-SR baselines.
\end{enumerate}

\vspace{-0.5em}
\section{Related Work}
\label{sec:related}

\noindent\textbf{EEG source reconstruction.}
The EEG inverse problem, recovering brain source activity from scalp measurements, is fundamentally ill-posed since the mapping from source to sensor space is many-to-one~\citep{grech2008review, hallez2007review}. Like many sampling scenarios, it becomes more severe with fewer or sparser samples.
Classical methods such as minimum norm estimation (MNE)~\citep{dale2000dynamic, hamalainen1994interpreting} and standardized low-resolution brain electromagnetic tomography (sLORETA)~\citep{pascual2002standardized} impose regularization priors (e.g., $\ell_2$ smoothness) to select among infinitely many solutions.
These methods require an explicit forward model (leadfield matrix) derived from structural MRI and head tissue conductivity models\citep{hamalainen1994interpreting,michel2019eeg}.
Recent work has explored data-driven or deep learning–based EEG source imaging that learns the source-to-sensor mapping end-to-end rather than relying solely on a fixed leadfield~\citep{huang2022electromagnetic,jiao2024multi,sun2022deep}. Our approach follows this trend but differs by parameterizing sources as explicit anisotropic 4D Gaussians on an anatomically anchored brain grid and embedding the forward operator as a differentiable Gaussian rendering step.

\noindent\textbf{EEG super-resolution.}
Deep learning approaches to EEG channel interpolation include convolutional and adversarial architectures that map LD to HD signals directly~\citep{corley2018deep, liu2023assigning, perez2025eeg,saba2020eeg}.
The Step-Aware Residual Guide~\citep{liu2025step} introduces residual learning with temporal step-aware conditioning, achieving strong performance with a lightweight CNN backbone by exploiting temporal structure.
While effective, these methods do not model the 3D geometry of the brain and electrodes, limiting physical interpretability.
Our method explicitly models sources in 3D brain space and renders to arbitrary electrode configurations.

\noindent\textbf{Neural fields and Gaussian splatting.}
Neural Radiance Fields (NeRF)~\citep{mildenhall2021nerf} represent scenes as continuous functions mapping spatial coordinates to radiance, using sinusoidal positional encodings~\citep{tancik2020fourier, vaswani2017attention} to capture high-frequency detail.
3D Gaussian Splatting~\citep{kerbl20233d} replaces the MLP with an explicit set of anisotropic 3D Gaussians (each with a full $3 \times 3$ covariance matrix), achieving real-time rendering.
4D extensions incorporate temporal dynamics: Wu et al.~\citep{wu20244d} combine 3D Gaussians with deformation fields, while 4D-Rotor Gaussian Splatting~\citep{duan20244d} uses native $4 \times 4$ spatiotemporal covariance matrices.

\noindent\textbf{Neural representations for EEG.}
Neural Brain Fields (NBF)~\citep{kedem2025neural} draws on the NeRF analogy for EEG, training a per-sample implicit neural representation to generate signals at unseen electrode positions.
While NBF demonstrates the potential of neural fields for EEG interpolation, it uses a black-box MLP without explicit source modeling or Gaussian parameterization.
Our approach extends this line of work by introducing explicit, interpretable Gaussian source components with full 4D covariance, combined with a physics-informed forward model.

\vspace{-0.5em}

\section{Method}
\label{sec:method}

\vspace{-0.5em}
\subsection{Problem formulation}
\label{sec:method:problem}

Let $\mathbf{X}^{\text{HD}} \in \mathbb{R}^{M \times T}$ denote an HD-EEG recording with $M$ electrodes and $T$ time steps, and let $\mathbf{X}^{\text{LD}} \in \mathbb{R}^{m \times T}$ ($m < M$) be a low-density subset obtained by selecting $m$ of the $M$ channels.
The super-resolution factor is $r = M / m$.
Our goal is to learn a mapping $f_\theta: \mathbb{R}^{m \times T} \to \mathbb{R}^{M \times T}$ that reconstructs $\hat{\mathbf{X}}^{\text{HD}} \approx \mathbf{X}^{\text{HD}}$ from $\mathbf{X}^{\text{LD}}$ alone.

Rather than learning this mapping directly in electrode space, we decompose it into two stages:
(1)~infer a latent 4D source field $\mathbf{S}(\mathbf{x}, t)$ for $\mathbf{x} \in \mathbb{R}^3$ and $t \in \{1, \ldots, T\}$, conditioned on the LD input; and
(2)~render the HD-EEG via a differentiable forward model that maps $\mathbf{S}$ to scalp potentials.

Our explicit Gaussian representation of $\mathbf{S}$ enables rendering to arbitrary HD electrode montages of any size, without retraining.

\vspace{-0.5em}
\subsection{Forward model: anisotropic 4D Gaussian field rendering}
\label{sec:method:forward}
Gaussian source components are anchored on a discrete brain grid $\mathcal{G} = \{\mathbf{g}_i\}_{i=1}^{N}$. We generate candidate points on a uniform cubic lattice spanning $[-90, 90]~\text{mm}$ in each spatial dimension at resolution $R$ (default $R = 12$) and retain only those within a sphere of radius $r_{\text{brain}} = 90~\text{mm}$:
\begin{equation}
    \mathcal{G} = \{ \mathbf{g}_i : \|\mathbf{g}_i\| \leq r_{\text{brain}} \}.
    \label{eq:braingrid}
\end{equation}
This yields $N \approx 826$ (at $R=12$) grid points, constraining sources to the anatomically relevant brain volume while keeping $N \ll M$ for efficiency.

At each grid point $\mathbf{g}_i$, we place $G$ Gaussian components (default $G = 3$), for a total of $N_{\text{total}} = N \cdot G$ components. Importantly, seeking a realistic setup, we assume time-dependent activation and thus allow time-modulated Gaussian components.
The predicted EEG signal at electrode $j$ and time $t$ is computed as a sum of Gaussian field contributions from all source components:
\begin{equation}
  \hat{X}_{j,t} = \sum_{n=1}^{N_{\text{total}}} a_n(t) \cdot \exp\!\left( -\frac{1}{2}\, \mathbf{d}_{j,n,t}^\top \boldsymbol{\Sigma}_n^{-1} \mathbf{d}_{j,n,t} \right),
  \label{eq:forward}
\end{equation}
where $a_n(t)$ is the modulated amplitude of component $n$ at time $t$ (Section~\ref{sec:method:ld}), $\mathbf{d}_{j,n,t} = [\mathbf{e}_j - \mathbf{g}_n;\, t_{\text{norm}} - \mu_n^{(t)}] \in \mathbb{R}^4$ is the 4D displacement vector between electrode $j$ and component $n$ at time $t$, and $\boldsymbol{\Sigma}_n^{-1}$ is the $4 \times 4$ precision matrix of the $n$-th Gaussian.

This formulation is analogous to Gaussian splatting in computer vision~\citep{kerbl20233d}: each source ``splats'' its contribution to nearby electrodes with a weight that decays as an anisotropic Gaussian function of the 4D space-time distance.
The full precision matrix enables each component to have different spatial spreads along different directions (anisotropy) and to couple spatial and temporal dimensions, so that a source's effective spatial extent can change over time.

\vspace{-0.5em}
\subsection{Mixture of anisotropic 4D Gaussians}
\label{sec:method:mixture}

The key innovation of EMAG is twofold: (1) placing \emph{multiple} Gaussian components at each grid point to form a mixture, enabling complex multi-modal source patterns, and (2) parameterizing each component with a full $4 \times 4$ precision matrix, enabling anisotropic spatial fields and explicit space-time coupling.

\noindent\textbf{Cholesky parameterization.}
A precision matrix must be symmetric positive-definite (SPD) to define a valid Gaussian shape.
We ensure this by construction via the Cholesky decomposition: we parameterize $\boldsymbol{\Sigma}_n^{-1}$ as $\mathbf{L}_n \mathbf{L}_n^\top$, where $\mathbf{L}_n$ is a lower-triangular matrix with positive diagonal entries.
Since any product $\mathbf{L} \mathbf{L}^\top$ is automatically SPD, the network can learn unconstrained values for $\mathbf{L}$ and the result is always a valid precision matrix:
\begin{equation}
  \boldsymbol{\Sigma}_n^{-1} = \mathbf{L}_n \mathbf{L}_n^\top, \quad
  \mathbf{L}_n = \begin{bmatrix}
    d_0 & 0 & 0 & 0 \\
    c_0 & d_1 & 0 & 0 \\
    c_1 & c_2 & d_2 & 0 \\
    c_3 & c_4 & c_5 & d_3
  \end{bmatrix},
  \label{eq:cholesky}
\end{equation}
where $d_i = \exp(\ell_i)$ are positive diagonal entries (stored as log-values $\ell_i$ for unconstrained optimization) and $c_j$ are free off-diagonal entries.
The diagonal entries control the precision (inverse of spread) along each axis: $d_0, d_1, d_2$ for the three spatial dimensions and $d_3$ for time.
The off-diagonal entries $c_0, \ldots, c_5$ control coupling between dimensions: for example, $c_3, c_4, c_5$ couple the spatial and temporal axes, allowing a Gaussian's spatial shape to vary with time.
This yields 10 learnable parameters per component (4 diagonal + 6 off-diagonal) and guarantees that $\boldsymbol{\Sigma}_n^{-1}$ is always SPD.

Each of the $N_{\text{total}}$ Gaussian components thus carries 12 learnable scalars: a signed \textbf{base amplitude} $w_n$ (overall source strength); a \textbf{temporal center} $\mu_n^{(t)}\in[0,1]$ (time of peak activation); the \textbf{Cholesky log-diagonal} $(\ell_0,\ell_1,\ell_2,\ell_3)$ controlling per-axis precision; and the \textbf{Cholesky off-diagonals} $(c_0,\ldots,c_5)$ encoding spatial anisotropy and space-time coupling.

\noindent\textbf{Efficient Mahalanobis decomposition.}
The exponent in Eq.~\ref{eq:forward} requires computing the Mahalanobis distance $\mathbf{d}^\top \boldsymbol{\Sigma}^{-1} \mathbf{d} = \|\mathbf{L}^\top \mathbf{d}\|^2$, which measures the distance from an electrode to a Gaussian center through the Gaussian's precision matrix.
Na\"ively computing this for every (electrode, component, timestep) triple requires materializing a $(B, T, M, N_{\text{total}}, 4)$ tensor, which is computationally expensive.
We exploit the fact that electrode and grid positions are fixed across time to decompose the distance into precomputable spatial terms and cheap per-timestep temporal terms:
\begin{equation}
  \|\mathbf{L}^\top \mathbf{d}\|^2 = \underbrace{A(\mathbf{e}_j, \mathbf{g}_n)}_{\text{spatial only}} + 2\tau \underbrace{C(\mathbf{e}_j, \mathbf{g}_n)}_{\text{coupling}} + \tau^2 \underbrace{D_n}_{\text{temporal only}},
  \label{eq:decomposition}
\end{equation}
where $\tau = t_{\text{norm}} - \mu_n^{(t)}$ and:
\begin{align}
  z_0 &= d_0 \Delta x + c_0 \Delta y + c_1 \Delta z, \quad
  z_1 = d_1 \Delta y + c_2 \Delta z, \quad
  z_2 = d_2 \Delta z, \label{eq:z_spatial} \\
  A &= z_0^2 + z_1^2 + z_2^2, \quad
  C = z_0 c_3 + z_1 c_4 + z_2 c_5, \quad
  D = c_3^2 + c_4^2 + c_5^2 + d_3^2. \label{eq:ACD}
\end{align}
The terms $A$, $C$, and $D$ are \textbf{precomputed once} per forward pass for all electrode–component pairs by combining the fixed spatial offsets $(\Delta x,\Delta y,\Delta z)$ with the current Cholesky parameters $(d_i,c_j)$ of each component.
Within a forward pass these quantities stay constant across time, while only the temporal displacement $\tau$ changes, so at each timestep we obtain the Mahalanobis distance by evaluating $\|\mathbf{L}^\top \mathbf{d}\|^2 = A(\mathbf{e}_j,\mathbf{g}_n) + 2\tau C(\mathbf{e}_j,\mathbf{g}_n) + \tau^2 D_n$, rather than recomputing a full 4D product for every time step.

\vspace{-0.5em}
\subsection{Low-density conditioning}
\label{sec:method:ld}

Rather than predicting source activity directly from space–time coordinates, we use the LD signals to drive a low-dimensional temporal code that modulates the amplitudes of Gaussians anchored on the brain grid, so that spatial structure is encoded in the grid while per-sample dynamics are driven only by the observations.
To condition the source field on the observed LD-EEG, we then need a mechanism that tells the Gaussian sources “how active to be’’ at each time step based on the sparse input.
We extract per-timestep features from $\mathbf{X}^{\text{LD}}$ using a lightweight 1D temporal encoder:
\begin{equation}
  \mathbf{h}_t = \text{TemporalEncoder}(\mathbf{X}^{\text{LD}}_{\cdot, t}) \in \mathbb{R}^{d_f},
  \label{eq:ld_encoder}
\end{equation}
where $d_f$ is the feature dimension.
This encoder reads the LD electrode values at time $t$ and compresses them into a feature vector.
A modulation network then maps this feature into a per-grid-point amplitude adjustment:
\begin{equation}
  \Delta w_i(t) = \text{MLP}_{\text{mod}}(\mathbf{h}_t)_i, \quad i = 1, \ldots, N.
  \label{eq:modulation}
\end{equation}
The modulation is shared across all $G$ Gaussians at the same grid point, preserving the structural relationship between co-located components.
The final amplitude for the $g$-th Gaussian at grid point $i$ at time $t$ is:
\begin{equation}
  a_{i,g}(t) = w_{i,g} + \Delta w_i(t).
  \label{eq:amplitude}
\end{equation}
In other words, each Gaussian has a fixed base amplitude $w_{i,g}$ (learned during training) plus a time-varying adjustment $\Delta w_i(t)$ driven by the LD input.
This separates the \emph{static} source layout (base amplitudes, spatial-temporal covariance) from the \emph{dynamic} modulation driven by the observed signal, analogous to how 4D Gaussian splatting separates static geometry from per-frame deformations~\citep{wu20244d}.

\vspace{-0.5em}
\subsection{Training objective}
\label{sec:method:training}

The model is trained end-to-end by minimizing the mean squared error between predicted and ground-truth HD-EEG, plus a small $L_2$ penalty on the source covariance and amplitude parameters that keeps the precision matrices well-conditioned:
\begin{equation}
  \mathcal{L} = \frac{1}{M \cdot T} \sum_{j=1}^{M} \sum_{t=1}^{T} \bigl(\hat{X}_{j,t} - X_{j,t}^{\text{HD}}\bigr)^2
  + \lambda \!\!\sum_{\theta \in \Theta_{\text{reg}}}\!\! \frac{1}{|\theta|}\|\theta\|_2^2,
  \label{eq:loss}
\end{equation}
where $\Theta_{\text{reg}} = \{\boldsymbol{\ell}_n, \mathbf{c}_n, w_n\}_n$ collects the Cholesky log-diagonal, off-diagonal, and base-amplitude parameters of every Gaussian, and $\lambda = 2\times 10^{-4}$.
We optimize with Adam~\citep{kingma2014adam} (learning rate $10^{-3}$, $\beta_1{=}0.9$, $\beta_2{=}0.999$), gradient clipping at $\|\nabla\|_2 \le 1$, and early stopping on validation loss (patience $20$ epochs, minimum improvement $10^{-4}$).

\vspace{-0.5em}
\section{Experimental evaluations}
\label{sec:experiments}
\vspace{-0.5em}
\subsection{Datasets}
\label{sec:experiments:dataset}

We evaluate EMAG on three public EEG benchmarks that span different experimental paradigms (preprocessing details in App.~\ref{app:datasets}):

\noindent\textbf{Localize-MI.} The Localize-MI dataset~\citep{mikulan2020simultaneous} provides 256-channel HD-EEG recordings from 7 human subjects during intracerebral electrical stimulation. Intracranial electrodes are used for stimulation only (not measurement); the scalp HD-EEG captures the brain's response. The known stimulation sites provide ground-truth source locations, which we use for post-hoc validation of learned source configurations (Section~\ref{sec:discussion}). Each subject has approximately 74 stimulation epochs at a sampling rate of 8~kHz. Given $M = 256$ HD electrodes, we construct LD subsets by uniformly subsampling channels at factors $r \in \{2, 4, 8, 16\}$, yielding $m \in \{128, 64, 32, 16\}$ LD electrodes respectively.

\noindent\textbf{SEED.} The SEED dataset~\citep{duan2013differential, zheng2015investigating} contains 62-channel EEG (international 10--20 montage, 200~Hz) recordings from 15 subjects during emotion elicitation via film clips, with three emotion classes (positive, neutral, negative). We use the preprocessed band-passed (0--75~Hz) signals released with the dataset. Each trial is segmented into non-overlapping 4~s windows (800 samples). LD subsets are constructed by uniformly subsampling channels at factors $r \in \{2, 4, 8\}$, yielding $m \in \{31, 15, 7\}$ LD electrodes.

\noindent\textbf{SEED-IV.} The SEED-IV dataset~\citep{zheng2018emotionmeter} extends SEED to four emotion classes (happy, sad, fear, neutral), with the same 62-channel 200~Hz montage across 15 subjects. SR setup, train/test split, and channel-subsampling protocol are identical to SEED ($r \in \{2,4,8\}$, $m \in \{31,15,7\}$).

For each dataset we use a per-subject 80/20 trial split and report metrics averaged across all of it's subjects; LD subset remains fixed per subject across sessions.

\vspace{-0.5em}
\subsection{Main results}
\label{sec:experiments:results}
\begin{table*}[]
  \caption{EEG super-resolution results across three datasets: \textbf{Localize-MI} (256 channels), \textbf{SEED}, and \textbf{SEED-IV} (62 channels). Best in \textbf{bold}, second-best \underline{underlined}. Metrics: NMSE ($\downarrow$), PCC ($\uparrow$), SNR ($\uparrow$, dB). All baseline results are from \citep{liu2025step}.}
  \label{tab:eeg_results}
  \centering
  \footnotesize
  \resizebox{\textwidth}{!}{
  \begin{tabular}{ll cccc ccc ccc}
    \toprule
    \multirow{2}{*}{Model} & \multirow{2}{*}{Metric} & \multicolumn{4}{c}{Localize-MI (256 ch.)} & \multicolumn{3}{c}{SEED (62 ch.)} & \multicolumn{3}{c}{SEED-IV (62 ch.)} \\
    \cmidrule(lr){3-6} \cmidrule(lr){7-9} \cmidrule(lr){10-12}
    & & $2\times$ & $4\times$ & $8\times$ & $16\times$ & $2\times$ & $4\times$ & $8\times$ & $2\times$ & $4\times$ & $8\times$ \\
    \midrule
    \multirow{3}{*}{SaSDim}
      & NMSE & $0.2675$ & $0.3427$ & $0.4174$ & $0.4613$ & $0.4399$ & $0.6234$ & $0.7767$ & $0.3633$ & $0.5543$ & $0.7122$ \\
      & PCC  & $0.8194$ & $0.7246$ & $0.6926$ & $0.6476$ & $0.7341$ & $0.5649$ & $0.4349$ & $0.7249$ & $0.6211$ & $0.5009$ \\
      & SNR  & $5.7443$ & $4.3796$ & $3.5549$ & $2.7678$ & $4.1154$ & $2.2940$ & $1.1349$ & $4.5940$ & $2.6004$ & $1.6211$ \\
    \midrule
    \multirow{3}{*}{SADI}
      & NMSE & $0.2637$ & $0.3442$ & $0.4164$ & $0.4566$ & $0.4439$ & $0.6049$ & $0.8106$ & $0.3557$ & $0.5349$ & $0.6844$ \\
      & PCC  & $0.8243$ & $0.7391$ & $0.6944$ & $0.6554$ & $0.7234$ & $0.5819$ & $0.4064$ & $0.7624$ & $0.6293$ & $0.5243$ \\
      & SNR  & $5.7511$ & $4.3724$ & $3.5498$ & $2.8942$ & $4.2419$ & $2.5160$ & $1.0137$ & $4.7093$ & $2.6044$ & $1.6610$ \\
    \midrule
    \multirow{3}{*}{RDPI}
      & NMSE & $0.2561$ & $0.3562$ & $0.4076$ & $0.4531$ & $0.4064$ & $0.6134$ & $0.7916$ & $0.3491$ & $0.5416$ & $0.6915$ \\
      & PCC  & $0.8246$ & $0.7396$ & $0.7062$ & $0.6549$ & $0.7416$ & $0.5716$ & $0.4216$ & $0.7861$ & $0.6316$ & $0.5164$ \\
      & SNR  & $5.7311$ & $4.3966$ & $3.5643$ & $2.7731$ & $4.2619$ & $2.3160$ & $1.0316$ & $4.7190$ & $2.6194$ & $1.6492$ \\
    \midrule
    \multirow{3}{*}{DDPMEEG}
      & NMSE & $0.2046$ & $0.3108$ & $0.3554$ & $0.4076$ & $0.4916$ & $0.7319$ & $0.8634$ & $0.5136$ & $0.6513$ & $0.7916$ \\
      & PCC  & $0.8516$ & $0.8163$ & $0.7306$ & $0.6739$ & $0.6941$ & $0.5134$ & $0.3419$ & $0.7346$ & $0.5316$ & $0.4305$ \\
      & SNR  & $6.2151$ & $5.5126$ & $3.9891$ & $3.2715$ & $4.1943$ & $1.5391$ & $0.9431$ & $4.4165$ & $2.1064$ & $1.6105$ \\
    \midrule
    \multirow{3}{*}{ESTformer}
      & NMSE & $0.2721$ & $0.3578$ & $0.4466$ & $0.4837$ & $0.3288$ & $0.3483$ & $0.4149$ & $0.3448$ & $0.3911$ & $0.5125$ \\
      & PCC  & $0.8061$ & $0.7205$ & $0.6867$ & $0.6319$ & $0.8368$ & $0.8012$ & $0.7670$ & $0.8106$ & $0.7822$ & $0.7048$ \\
      & SNR  & $5.5403$ & $3.8671$ & $3.3023$ & $2.5671$ & $5.0560$ & $4.5838$ & $3.8871$ & $4.7535$ & $4.1933$ & $2.9821$ \\
    \midrule
    \multirow{3}{*}{STAD}
      & NMSE & $0.1902$ & $0.3067$ & $0.3649$ & $0.4106$ & $0.4319$ & $0.6913$ & $0.8671$ & $0.3819$ & $0.6713$ & $0.7193$ \\
      & PCC  & $0.8635$ & $0.8194$ & $0.7216$ & $0.6694$ & $0.7136$ & $0.4946$ & $0.3441$ & $0.7316$ & $0.5219$ & $0.4319$ \\
      & SNR  & $7.2591$ & $5.5234$ & $3.8715$ & $3.2642$ & $4.1364$ & $1.4349$ & $0.9134$ & $4.4930$ & $2.0492$ & $1.6193$ \\
    \midrule
    \multirow{3}{*}{SRGDiff}
      & NMSE & \underline{0.1449} & \underline{0.2384} & \underline{0.2957} & \underline{0.3457} & \underline{0.1632} & \underline{0.2977} & \textbf{0.3494} & \underline{0.1663} & \underline{0.2115} & \textbf{0.2603} \\
      & PCC  & \underline{0.9213} & \underline{0.8854} & \underline{0.8323} & \underline{0.7322} & \underline{0.9102} & \underline{0.8445} & \textbf{0.8167} & \underline{0.9113} & \underline{0.8846} & \textbf{0.8210} \\
      & SNR  & \underline{8.3755} & \underline{6.3617} & \underline{5.2249} & \underline{4.0197} & \underline{7.8413} & \underline{5.2606} & \textbf{4.5912} & \underline{7.8660} & \underline{6.6402} & \textbf{6.0346} \\
    \midrule
    \multirow{3}{*}{EMAG (ours)}
      & NMSE & \textbf{0.0856} & \textbf{0.1040} & \textbf{0.1396} & \textbf{0.2139} & \textbf{0.1545} & \textbf{0.2776} & \underline{0.4027} & \textbf{0.1153} & \textbf{0.2027} & \underline{0.3358} \\
      & PCC  & \textbf{0.9561} & \textbf{0.9464} & \textbf{0.9274} & \textbf{0.8863} & \textbf{0.9194} & \textbf{0.8496} & \underline{0.7719} & \textbf{0.9404} & \textbf{0.8926} & \underline{0.8128} \\
      & SNR  & \textbf{10.8996} & \textbf{10.0156} & \textbf{8.6731} & \textbf{6.7959} & \textbf{8.2044} & \textbf{5.6243} & \underline{3.9989} & \textbf{9.5780} & \textbf{7.0593} & \underline{4.7712} \\
    \bottomrule
  \end{tabular}
  }
\end{table*}

We compare EMAG against SRGDiff~\citep{liu2025step}---the current SOTA for deep EEG super-resolution---plus all baselines reported there: SaSDim~\citep{zhang2023sasdim}, SADI~\citep{islam2025self}, RDPI~\citep{liu2025rdpi}, DDPMEEG~\citep{vetter2024generating}, ESTformer~\citep{li2023estformer}, and STAD~\citep{wang2025generative}. Alternative 4D Gaussian parameterizations are reported in Supp.~Sec.~\ref{app:ablations}. Per-subject numbers and standard deviations are tabulated in Supp.~Sec.~\ref{app:qualitative_examples}.

\begin{figure}[t]
  \centering
  \caption{Qualitative reconstruction (SEED sub-01, $r{=}4$, peak-GFP timestep, PCC$=0.99$)
  Additional waveform-domain examples in App.~\ref{app:qualitative_examples:waveforms}.}
  \includegraphics[width=\linewidth]{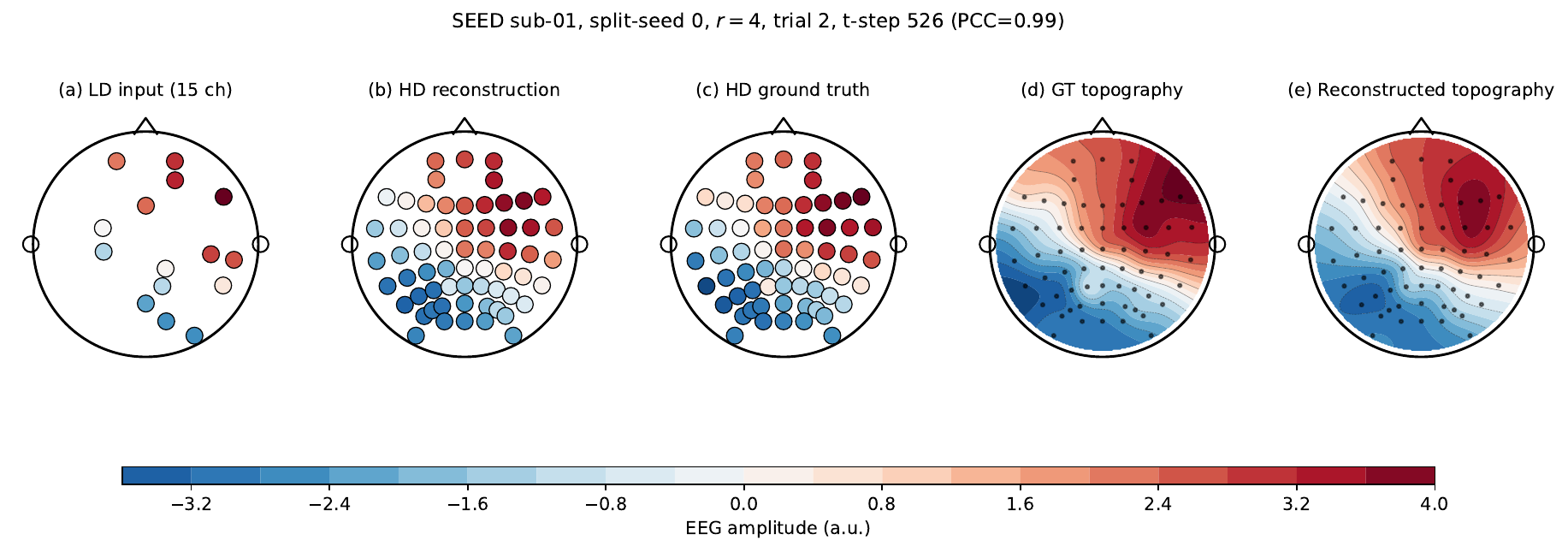}
  \label{fig:recon_topomap}
\end{figure}

Table~\ref{tab:eeg_results} reports reconstruction quality on Localize-MI (256 ch., 7 subjects), SEED, and SEED-IV (62 ch., 15 subjects each) across all SR factors. EMAG leads on Localize-MI at every $r$, cutting NMSE by $41$\,--\,$62\%$ and improving SNR by $2.5$\,--\,$4.4$\,dB relative to the strongest prior baseline (SRGDiff), while keeping PCC above $0.88$ even at $r{=}16$. On SEED and SEED-IV it dominates at $r{\in}\{2,4\}$ across all three metrics; at $r{=}8$ it trails SRGDiff in NMSE/SNR while remaining second-best in PCC, an outcome consistent with information-theoretic limits of $8\times$ on a $62$-channel cap, qualitative reconstruction example in Fig.~\ref{fig:recon_topomap}.

Three observations clarify these trends. \emph{(i)}~The advantage is largest on Localize-MI, whose dense $256$-channel sEEG montage reward EMAG's anisotropic 4D source representation; degradation with $r$ is also slowest there ($\Delta\mathrm{PCC}{=}0.07$ between $r{=}2$ and $r{=}16$, vs.\ $0.13$\,--\,$0.15$ for the next-best method), indicating stronger robustness to extreme decimation than electrode-space diffusion. \emph{(ii)}~On $62$-channel scalp data, the EMAG–SRGDiff gap shrinks with $r$ as the inverse problem becomes under-determined, aligning with the LD conditioning ablation (Sec.~\ref{sec:ablation}) that identifies input spatial resolution as the limiting factor beyond the cap’s bandwidth. \emph{(iii)}~EMAG achieves this with ${\sim}76$k parameters per subject—about $30\times$ fewer than SRGDiff (${\sim}2.3$M)—and trains in under $30$ minutes on a single GPU, supporting its role as a lightweight \emph{personalized} model (cross-subject transfer in Sec.~\ref{sec:experiments:crosssubj}).

\vspace{-0.5em}
\subsection{Cross-subject generalization}
\label{sec:experiments:crosssubj}

EMAG is by design a \emph{per-subject} model: the 4D Gaussian field is fit on each subject. To quantify how much of EMAG's signal is subject-specific, for each (dataset, $r$) we apply every subject-trained model to every other subject's LD recordings and report the mean off-diagonal score of the resulting $S\times S$ transfer matrix. As a non-learned reference we upsample the LD montage to the HD grid via spherical splines (\textit{LD-spline}). Cross-subject EMAG never matches its within-subject score ($\Delta\mathrm{PCC}\!\approx\!0.31$\,--\,$0.39$) yet remains markedly better than LD-spline at every $r$ on SEED/SEED-IV (e.g.\ at $r{=}2$: cross PCC\,$0.565$ vs LD-spline $0.469$). This indicates that the Gaussian field encodes geometric priors reusable across subjects while per-subject training captures the remaining individual cortical layout. We therefore position EMAG as a lightweight \emph{personalized} SR module---trainable from a small per-subject calibration set (${\sim}76$k parameters)---rather than a universal cross-subject network. Full $S\times S$ heat-maps, the LD-spline construction, the Localize-MI exclusion rationale, and per-seed numbers are in App.~\ref{app:crosssubj} (Table~\ref{tab:supp_crosssubj_full}, Fig.~\ref{fig:crosssubj_heatmaps}).

\vspace{-0.5em}
\subsection{Ablation studies}
\label{sec:ablation}

We conduct a systematic ablation study to validate each core design decision in EMAG, evaluated on Localize-MI and SEED at $4\times$ super-resolution. Full results are reported in Table~\ref{tab:combined_ablation} and a detailed per-ablation discussion is provided in Appendix~\ref{app:ablations}.

\noindent\textbf{4D Gaussian parameterization.}
Reducing the precision matrix from full $4{\times}4$ to spatial-only ($3{\times}3$), diagonal, and isotropic-scalar variants degrades performance monotonically (isotropic: $+19.0\%$ NMSE on Localize-MI, $+22.1\%$ on SEED). Zeroing only the spatial off-diagonals $c_0,c_1,c_2$ or only the spatio-temporal $c_3,c_4,c_5$ produces comparable degradation, indicating the two coupling types are complementary and jointly contribute to optimal performance.

\noindent\textbf{Brain grid and source placement.}
Without the spherical grid anchor, free-floating components migrate to the scalp and act as shortcut interpolators, so the constraint is a geometric regularizer rather than a convenience. Restricting the grid to a thin cortical shell costs $+116.2\%$ NMSE on SEED, challenging the assumption that emotion-related EEG is purely cortical.

\noindent\textbf{Low-density conditioning.}
Removing the TemporalEncoder collapses NMSE to $\approx 1.0$ on both datasets, and replacing per-grid-point modulation with a global scalar costs $+434\%$ NMSE on Localize-MI yet only $+7.6\%$ on SEED, reflecting the spatial focality of evoked responses. Pre-interpolating the LD input yields no gain, indicating the Gaussian forward model already supplies the spatial regularization.

\noindent\textbf{Physics-informed forward model.}
A direct electrode-space MLP raises Localize-MI NMSE by $+113.7\%$ but SEED by only $+2.0\%$, again tracking spatial focality. A learned linear source-to-electrode projection matches the MLP on Localize-MI yet collapses on SEED ($+94.0\%$), exposing its lack of a distance-decay prior; the fixed three-shell leadfield is the weakest variant on both datasets, unable to adapt to subject anatomy.

\begin{table*}[t]
  \centering
  \caption{Ablation results on \textbf{Localize-MI} and \textbf{SEED} at $4\times$ super-resolution.}
  \label{tab:combined_ablation}
  \footnotesize
  \resizebox{\textwidth}{!}{
  \begin{tabular}{ll ccc ccc}
    \toprule
    \multirow{2}{*}{\textbf{Category}} & \multirow{2}{*}{\textbf{Variant}} & \multicolumn{3}{c}{\textbf{Localize-MI} (mean across 7 subjects)} & \multicolumn{3}{c}{\textbf{SEED} (mean across 15 subjects)} \\
    \cmidrule(lr){3-5} \cmidrule(lr){6-8}
    & & \textbf{NMSE} $\downarrow$ & \textbf{PCC} $\uparrow$ & \textbf{SNR} $\uparrow$ & \textbf{NMSE} $\downarrow$ & \textbf{PCC} $\uparrow$ & \textbf{SNR} $\uparrow$ \\
    \midrule
    Proposed Model & \textbf{EMAG (full)} & \textbf{0.1013} & \textbf{0.9478} & \textbf{10.1427} & \textbf{0.2776} & \textbf{0.8496} & \textbf{5.6243} \\
    \midrule
    \multirow{5}{*}{\shortstack[l]{4D Gaussian\\Parameterization}}
      & w/o temporal axis ($3\times3$)   & 0.1095 & 0.9435 & 9.7797 & 0.3036 & 0.8341 & 5.2206 \\
      & Diagonal precision (no coupling) & 0.1094 & 0.9436 & 9.7841 & 0.2980 & 0.8375 & 5.3035 \\
      & Only spatial coupling            & 0.1096 & 0.9435 & 9.7791 & 0.2981 & 0.8374 & 5.3045 \\
      & Only spatio-temporal coupling    & 0.1095 & 0.9435 & 9.7810 & 0.2987 & 0.8371 & 5.2933 \\
      & Isotropic scalar $\sigma$        & 0.1205 & 0.9376 & 9.3355 & 0.3388 & 0.8128 & 4.7357 \\
    \midrule
    \multirow{2}{*}{\shortstack[l]{Brain Grid \&\\Source Placement}}
      & Free-floating centers            & 0.1095 & 0.9435 & 9.7817 & 0.2976 & 0.8377 & 5.3094 \\
      & Surface-only grid                & 0.1370 & 0.9288 & 8.7703 & 0.6002 & 0.6316 & 2.2260 \\
    \midrule
    \multirow{3}{*}{\shortstack[l]{LD\\Conditioning}}
      & No conditioning (static)         & 0.9995 & 0.0112 & 0.0023 & 1.0000 & 0.0034 & 0.0001 \\
      & Global amplitude modulation      & 0.5414 & 0.6762 & 2.6841 & 0.2986 & 0.8371 & 5.2960 \\
      & Pre-interpolated LD input        & 0.1197 & 0.9381 & 9.3760 & 0.3014 & 0.8354 & 5.2520 \\
    \midrule
    \multirow{3}{*}{\shortstack[l]{Forward\\Model}}
      & Direct electrode mapping         & 0.2165 & 0.8845 & 6.9413 & 0.2831 & 0.8463 & 5.5331 \\
      & Learned linear projection        & 0.2122 & 0.8873 & 6.8383 & 0.5384 & 0.6791 & 2.6970 \\
      & Fixed leadfield matrix           & 0.2573 & 0.8614 & 5.9793 & 0.5955 & 0.6356 & 2.2600 \\
    \bottomrule
  \end{tabular}
  }
\end{table*}

\vspace{-0.5em}
\section{Which channels should we keep? An SR-fidelity study of electrode subsets}
\label{sec:elecsub}

Throughout the paper we have treated the low-density (LD) montage as a uniformly-random $K$-subset of the HD electrodes. In practice, sparse-EEG hardware imposes a \emph{fixed} electrode layout, and a long line of EEG channel-selection work has argued that not all electrodes are equally informative for downstream tasks \citep{abdullah2022eeg,kim2022miniaturization}.
We ask the \emph{reverse} question that is natural for super-resolution: when the goal is to \emph{reconstruct} the full montage from $K$ retained channels, do these anatomically-motivated subsets remain a good choice, or does the inverse problem favour layouts with different geometry?

For that manner, we define ten named subsets at three sparsification levels matching the SR factors used in the main benchmark (Visualization in figure~\ref{fig:electrode_layouts}, and full channel lists in Table~\ref{tab:elecsub:defs}, both in Appendix~\ref{sec:elecsub:supp}): SR$\times$2 ($K{=}31$): \textsc{Hemi-Left}, \textsc{Hemi-Right} (left/right halves plus the four midline channels Fz, Cz, Pz, Oz); SR$\times$4 ($K{=}15$): \textsc{V15} (rim-style superset of the ten circumference channels of \citet{valderrama2025identifying}), \textsc{FT15} (fronto-temporal), \textsc{INT15} (central / midline control); SR$\times$8 ($K{=}7$): four hemispheric rim slices (\textsc{VL7}, \textsc{VR7}, \textsc{VU7}, \textsc{VLw7}) and an \textsc{INT7} (central) control.

\noindent\textbf{Results.}
Two findings stand out. \emph{(i) Uniform spatial coverage dominates anatomical priors.} Across \emph{every} tested (dataset, $r$) cell, uniform \textsc{Random} selection consistently achieves the lowest NMSE and highest PCC/SNR among the named subsets evaluated. The gap grows with the sparsification factor: at SR$\times$2 the named hemispheric subsets cost roughly $+0.16$--$0.17$ NMSE on both datasets, at SR$\times$4 roughly $+0.10$--$0.14$, and at SR$\times$8 roughly $+0.10$--$0.15$, and the seed-to-seed standard deviation of the named subsets ($\le 0.06$ NMSE) is an order of magnitude smaller than the gap to \textsc{Random}(Table~\ref{tab:elecsub:fidelity_main}). We attribute this to the geometry of the inverse problem: a uniform random $K$-subset spreads observations isotropically over the scalp, while Anatomically-clustered subsets, by contrast, leave large portions of the scalp with no nearby observed electrode, and EMAG must extrapolate over those gaps.

\emph{(ii) Differences among named subsets are small and not monotonic in the rim-vs-interior axis.} At SR$\times$4, the rim-style \textsc{V15} and \textsc{FT15} subsets slightly outperform the interior control \textsc{INT15} on both datasets (NMSE gap $\approx 0.03$). At SR$\times$8, however, the central \textsc{INT7} is competitive or better than every rim slice; at this regime the absolute size $K{=}7$ is small enough that any clustered subset under-samples the rest of the head. The two SR$\times$2 hemispheric subsets are nearly indistinguishable (Table~\ref{tab:elecsub:fidelity_main}), in line with the small spatial sensitivity of the reconstructor at low SR factors.

\begin{table}[t]
  \caption{EEG super-resolution fidelity on \textbf{SEED} and \textbf{SEED-IV} for each named electrode subset versus the uniform-random baseline. Metrics: NMSE ($\downarrow$), PCC ($\uparrow$), SNR in dB ($\uparrow$). Per-cell $\Delta_{\text{rand}}$ numbers for all three metrics are reported in Table~\ref{tab:elecsub:fidelity_full} of Appendix~\ref{sec:elecsub:supp}.}
  \label{tab:elecsub:fidelity_main}
  \centering
  \footnotesize
  \setlength{\tabcolsep}{4pt}
  \resizebox{\textwidth}{!}{%
  \begin{tabular}{l l ccc ccc}
    \toprule
    & & \multicolumn{3}{c}{\textbf{SEED} (62 ch.)}
      & \multicolumn{3}{c}{\textbf{SEED-IV} (62 ch.)} \\
    \cmidrule(lr){3-5}\cmidrule(lr){6-8}
    SR & Subset
      & NMSE $\downarrow$ & PCC $\uparrow$ & SNR (dB) $\uparrow$
      & NMSE $\downarrow$ & PCC $\uparrow$ & SNR (dB) $\uparrow$ \\
    \midrule
    \multirow{3}{*}{$\times 2$}
      & \textsc{Random}       & $\mathbf{0.154{\scriptstyle\pm0.001}}$ & $\mathbf{0.919{\scriptstyle\pm0.001}}$ & $\mathbf{8.20{\scriptstyle\pm0.03}}$
                              & $\mathbf{0.115{\scriptstyle\pm0.001}}$ & $\mathbf{0.940{\scriptstyle\pm0.001}}$ & $\mathbf{9.58{\scriptstyle\pm0.03}}$ \\
      & \textsc{Hemi-Left}    & $0.312{\scriptstyle\pm0.058}$ & $0.829{\scriptstyle\pm0.035}$ & $5.15{\scriptstyle\pm0.77}$
                              & $0.285{\scriptstyle\pm0.030}$ & $0.845{\scriptstyle\pm0.018}$ & $5.50{\scriptstyle\pm0.45}$ \\
      & \textsc{Hemi-Right}   & $0.312{\scriptstyle\pm0.050}$ & $0.829{\scriptstyle\pm0.031}$ & $5.13{\scriptstyle\pm0.68}$
                              & $0.275{\scriptstyle\pm0.032}$ & $0.851{\scriptstyle\pm0.019}$ & $5.67{\scriptstyle\pm0.50}$ \\
    \midrule
    \multirow{4}{*}{$\times 4$}
      & \textsc{Random}       & $\mathbf{0.278{\scriptstyle\pm0.002}}$ & $\mathbf{0.850{\scriptstyle\pm0.001}}$ & $\mathbf{5.62{\scriptstyle\pm0.02}}$
                              & $\mathbf{0.203{\scriptstyle\pm0.001}}$ & $\mathbf{0.893{\scriptstyle\pm0.000}}$ & $\mathbf{7.06{\scriptstyle\pm0.02}}$ \\
      & \textsc{V15}          & $0.384{\scriptstyle\pm0.037}$ & $0.784{\scriptstyle\pm0.024}$ & $4.22{\scriptstyle\pm0.42}$
                              & $0.299{\scriptstyle\pm0.010}$ & $0.837{\scriptstyle\pm0.006}$ & $5.29{\scriptstyle\pm0.15}$ \\
      & \textsc{FT15}         & $0.385{\scriptstyle\pm0.038}$ & $0.783{\scriptstyle\pm0.024}$ & $4.19{\scriptstyle\pm0.42}$
                              & $0.311{\scriptstyle\pm0.020}$ & $0.829{\scriptstyle\pm0.012}$ & $5.12{\scriptstyle\pm0.28}$ \\
      & \textsc{INT15}        & $0.418{\scriptstyle\pm0.037}$ & $0.762{\scriptstyle\pm0.024}$ & $3.83{\scriptstyle\pm0.38}$
                              & $0.330{\scriptstyle\pm0.025}$ & $0.818{\scriptstyle\pm0.015}$ & $4.89{\scriptstyle\pm0.33}$ \\
    \midrule
    \multirow{6}{*}{$\times 8$}
      & \textsc{Random}       & $\mathbf{0.422{\scriptstyle\pm0.003}}$ & $\mathbf{0.760{\scriptstyle\pm0.002}}$ & $\mathbf{3.80{\scriptstyle\pm0.03}}$
                              & $\mathbf{0.336{\scriptstyle\pm0.001}}$ & $\mathbf{0.814{\scriptstyle\pm0.001}}$ & $\mathbf{4.80{\scriptstyle\pm0.01}}$ \\
      & \textsc{VL7}          & $0.540{\scriptstyle\pm0.017}$ & $0.677{\scriptstyle\pm0.013}$ & $2.70{\scriptstyle\pm0.14}$
                              & $0.501{\scriptstyle\pm0.001}$ & $0.706{\scriptstyle\pm0.001}$ & $3.02{\scriptstyle\pm0.01}$ \\
      & \textsc{VR7}          & $0.527{\scriptstyle\pm0.001}$ & $0.687{\scriptstyle\pm0.001}$ & $2.80{\scriptstyle\pm0.01}$
                              & $0.491{\scriptstyle\pm0.000}$ & $0.713{\scriptstyle\pm0.000}$ & $3.10{\scriptstyle\pm0.00}$ \\
      & \textsc{VU7}          & $0.561{\scriptstyle\pm0.002}$ & $0.662{\scriptstyle\pm0.002}$ & $2.53{\scriptstyle\pm0.02}$
                              & $0.470{\scriptstyle\pm0.001}$ & $0.727{\scriptstyle\pm0.001}$ & $3.32{\scriptstyle\pm0.01}$ \\
      & \textsc{VLw7}         & $0.572{\scriptstyle\pm0.004}$ & $0.654{\scriptstyle\pm0.003}$ & $2.44{\scriptstyle\pm0.03}$
                              & $0.469{\scriptstyle\pm0.005}$ & $0.727{\scriptstyle\pm0.003}$ & $3.34{\scriptstyle\pm0.05}$ \\
      & \textsc{INT7}         & $0.529{\scriptstyle\pm0.006}$ & $0.685{\scriptstyle\pm0.005}$ & $2.78{\scriptstyle\pm0.05}$
                              & $0.441{\scriptstyle\pm0.001}$ & $0.747{\scriptstyle\pm0.001}$ & $3.61{\scriptstyle\pm0.01}$ \\
    \bottomrule
  \end{tabular}}
\end{table}

\vspace{-0.5em}
\section{Discussion and Conclusion}
\label{sec:discussion}

\noindent\textbf{Why anisotropic 4D Gaussians work for EEG.}
Scalp EEG is a volume-conducted projection of cortical sources whose currents are oriented (perpendicular to the cortical sheet) and produce anisotropic scalp patterns that isotropic kernels cannot represent. The full $4{\times}4$ precision matrix captures this spatial anisotropy and couples it to the temporal evolution of source patterns in a single parameterization.

\noindent\textbf{Comparison with prior reconstruction methods.}
Unlike MNE~\citep{hamalainen1994interpreting} or sLORETA~\citep{pascual2002standardized}, EMAG learns its forward model jointly with the source parameters and needs no precomputed leadfield from structural MRI, so it applies even when individual anatomy is unavailable. Unlike electrode-space CNN/diffusion baselines (e.g.\ SRGDiff~\citep{liu2025step}) it also exposes the underlying source configuration, which we exploit in the explainability paragraph below.

\noindent\textbf{Model explainability.}
\label{sec:discussion:explainability}
The learned Gaussian configurations are directly interpretable: each component carries a position $\mathbf{g}_n$, a precision matrix $\boldsymbol{\Sigma}_n^{-1}$, and a signed amplitude $w_n$. After training, high-$|w_n|$ components cluster near the known stimulation sites in Localize-MI (App.~\ref{app:interp:sources}, Fig.~\ref{fig:gaussian_sources}); the full $4{\times}4$ covariance lets each component be classified as point-like, elongated (directed cortical current), or sheet-like (App.~\ref{app:interp:anisotropy}, Fig.~\ref{fig:anisotropy}; top-$K$ ellipsoids in Fig.~\ref{fig:source_validation_3d}); and the coupling terms $c_3,c_4,c_5$ let a Gaussian's spatial extent vary in time. This level of source-space readout is not available from CNN/diffusion SR baselines.

\noindent\textbf{When uniform sampling beats anatomy.}
Channel-selection recipes from the classifier-driven EEG literature do not transfer to super-resolution: anatomically-clustered subsets leave large scalp regions unobserved and a generative reconstructor pays a measurable fidelity cost (Sec.~\ref{sec:elecsub}, App.~\ref{sec:elecsub:app:discussion}). Layouts that maximise spatial coverage, which uniform random sampling approximates, are the better default for SR pipelines.

\noindent\textbf{Limitations.}
The current model is trained per-subject; a cross-subject or subject-adaptive variant would improve practical applicability. The spherical brain grid remains a simplification of the true brain geometry; future work could use subject-specific cortical meshes. Evaluation is currently focused on three datasets; broader validation on naturalistic EEG paradigms (e.g., resting state, BCI motor imagery) is warranted.

\textbf{EMAG} demonstrates that physics-informed 4D Gaussian splatting offers a practical, interpretable path for EEG super-resolution---lightweight personalized models that unlock HD insights from sparse clinical hardware. By bridging computer vision rendering techniques with EEG's ill-posed inverse problem, we hope EMAG opens new directions for neural field representations in neuroscience. All code and experiments will be made publicly available upon acceptance. Additionally, We provide an ethics and responsible-use discussion in App.~\ref{app:ethics}.


\bibliographystyle{plainnat}
\bibliography{references}

\appendix

\section{Implementation details}
\label{app:implementation}

\noindent\textbf{Brain grid.}
The brain domain spans $[-90, 90]~\text{mm}$ in each spatial dimension. The brain grid at resolution $R$ consists of all points on a uniform $R \times R \times R$ lattice that fall within a sphere of radius $90~\text{mm}$. For the default $R = 12$, the sphere grid yields $N \approx 826$ grid points. With $G = 3$ Gaussians per point, the default configuration has $N_{\text{total}} \approx 2{,}478$ components and ${\sim}76$K total learnable parameters.

\noindent\textbf{Per-component parameterization.}
Each Gaussian has 12 learnable parameters: 1 base amplitude, 1 temporal center $\mu^{(t)}$, 4 Cholesky diagonal entries (log-scale), and 6 Cholesky off-diagonal entries. Total per-Gaussian parameters: $12 \times N_{\text{total}} \approx 29{,}700$ for the default configuration. The LD encoder adds ${\sim}2$K parameters; the modulation MLP (mapping $\mathbb{R}^{32} \to \mathbb{R}^{64} \to \mathbb{R}^{N}$) adds ${\sim}55$K parameters at $N \approx 826$, dominating the total parameter budget of ${\sim}76$K.

\noindent\textbf{Computational efficiency.}
The Cholesky-based Mahalanobis decomposition (Eq.~\ref{eq:decomposition}) enables EMAG to use the full 4D covariance at minimal overhead over isotropic variants. The spatial terms $A$ and $C$ are precomputed once per forward pass, and only the scalar temporal displacement $\tau$ varies at each time step, making the per-timestep cost $\mathcal{O}(M\cdot N_{\text{total}})$.

\noindent\textbf{Initialization.}
Base amplitudes $w_n$ are initialized to zero. Temporal centers $\mu_n^{(t)}$ are drawn from $\text{Uniform}(0, 0.5)$ (normalized time). Cholesky diagonal entries are initialized to $\ell_i = \log(1/15)$ for spatial dimensions (corresponding to a spatial precision of $1/15~\text{mm}^{-1}$, i.e., a spatial spread of ${\sim}15~\text{mm}$) and $\ell_3 = \log(1/0.1)$ for the temporal dimension. Off-diagonal Cholesky entries are initialized to zero (no coupling at initialization).

\noindent\textbf{LD encoder.}
The temporal encoder is a 1D convolution over the LD channels at each time step, producing a $d_f = 32$-dimensional feature vector. The modulation MLP has one hidden layer with 64 units and ReLU activation, mapping from $\mathbb{R}^{d_f}$ to $\mathbb{R}^N$ (one value per grid point, shared across $G$ Gaussians).

\noindent\textbf{Training protocol.}
All models are trained for up to 100 epochs with early stopping (patience 20, minimum improvement $10^{-4}$) using Adam~\citep{kingma2014adam} (learning rate $10^{-3}$, $\beta_1 = 0.9$, $\beta_2 = 0.999$). Each epoch is processed in temporal chunks with gradient accumulation. Training uses a single NVIDIA RTX 6000 Ada (50.9~GB) GPU and completes in under 30 minutes per subject for the default configuration.

\noindent\textbf{Alternative parameterizations.}
The NeRF-Hybrid, Per-Voxel, and Isotropic Mixture 4D parameterizations share the same training protocol and LD encoder. Full hyperparameter details for each variant are provided in the supplementary material.

\noindent\textbf{Compute resources.}
All experiments ran on a single internal SLURM cluster (NVIDIA RTX6000 GPUs with $\sim$48\,GB VRAM, single-GPU allocation per job; 4 CPU cores, 24\,GB RAM). A single (subject, SR-factor) EMAG training run takes 30--120\,min wall-clock at $r{=}2$--$16$. The full benchmark (3 datasets $\times$ 3--4 SR factors $\times$ 3 seeds $\times$ 7--15 subjects) plus the ablation grid (Section~\ref{app:ablations}) consumed approximately 180 GPU-hours; preliminary and failed runs (hyperparameter search, debugging) add an estimated additional $1{-}2{\times}$.


\section{Datasets}
\label{app:datasets}

\noindent\textbf{Dataset preprocessing and split protocol.}
For Localize-MI we use the released averaged-reference 256-ch HD-EEG at 8\,kHz, retain the $\pm$\,200\,ms window around stimulation onset, and apply a per-subject $z$-score with statistics computed on the training split only. For SEED and SEED-IV we use the preprocessed 0–75\,Hz band-pass signals at 200\,Hz released with each dataset, segment into non-overlapping 4\,s windows (800 samples), and apply per-channel $z$-scoring with train-split statistics. The 80/20 train/test split is by trial within subject; the random split seed is fixed per run and we report mean$\pm$std across three seeds $s\in\{0,1,2\}$ unless otherwise noted. The LD electrode subset is sampled uniformly without replacement from the HD montage, with seed $=0$ throughout the main benchmark, and held fixed across all sessions of a subject.

\noindent\textbf{Datasets, licenses and access.}
\textbf{Localize-MI}~\citep{mikulan2020simultaneous}: released under \href{https://creativecommons.org/licenses/by/4.0/}{CC BY 4.0}; available from OpenNeuro (\texttt{ds003644}). \textbf{SEED}~\citep{zheng2015investigating} and \textbf{SEED-IV}~\citep{duan2013differential}: released under their original BCMI laboratory data-use agreement; available upon application from \href{https://bcmi.sjtu.edu.cn/home/seed/}{bcmi.sjtu.edu.cn/home/seed}. \textbf{Baselines.} SaSDim, SADI, RDPI, DDPMEEG, ESTformer, STAD and SRGDiff are used under their respective original licenses; we cite the source papers in Section~\ref{sec:experiments:results} and Section~\ref{sec:related} and release no modified weights.

\section{Cross-subject generalization: extended results}
\label{app:crosssubj}

This section gives the full protocol, per-(dataset, $r$) numbers with standard deviations, and additional discussion accompanying Section~\ref{sec:experiments:crosssubj}.

\noindent\textbf{Protocol.}
For every dataset $\mathcal{D}\in\{\textrm{SEED},\textrm{SEED-IV}\}$ and SR factor $r\in\{2,4,8\}$ we use the per-subject EMAG checkpoints already trained for the main results in Table~\ref{tab:eeg_results} (architecture: $R{=}12$, $G{=}3$, spherical brain grid, random LD selection, 3 split seeds). Let $S$ denote the number of subjects in $\mathcal{D}$ ($S{=}15$ for both SEED and SEED-IV). For each split seed $s$ we form the $S\times S$ transfer matrix $T^{(s)}$ where $T^{(s)}_{ij}$ is obtained by:
\begin{enumerate}\itemsep0pt
  \item loading the EMAG model trained on subject $i$ (split seed $s$);
  \item feeding it subject $j$'s LD test recordings (the same LD subset and
        test split that was used when evaluating subject $j$ in the main
        results, so $T^{(s)}_{jj}$ exactly reproduces the within-subject
        number);
  \item recording NMSE, PCC and SNR between the model's output and
        subject $j$'s HD ground truth.
\end{enumerate}
We report
\[
\textit{within}^{(s)} \;=\; \tfrac{1}{S}\sum_{i} T^{(s)}_{ii},
\qquad
\textit{cross}^{(s)} \;=\; \tfrac{1}{S(S-1)}\sum_{i\neq j} T^{(s)}_{ij},
\]
and average across seeds. Each non-diagonal cell $T^{(s)}_{ij}$ uses no information from subject $j$ during training: subject $i$'s 4D Gaussian field is applied verbatim to subject $j$'s LD signal, with subject $j$'s electrode positions used only to evaluate the rendered output (the LD-conditioning MLP is, by design, shared across grid points but not across subjects).

\noindent\textbf{LD-spline reference.}
As a model-free baseline we use spherical-spline interpolation: each LD sample (on the $K{=}n_{\rm hd}/r$ retained electrodes) is upsampled to the full HD montage by fitting a Legendre-polynomial spherical spline ($m{=}4$, $\lambda{=}10^{-5}$) per time step. The resulting HD prediction depends only on subject $j$'s own LD signal and electrode positions, so it is by construction a cross-subject method (no training subject involved); we therefore report a single LD-spline column rather than a transfer matrix.

\noindent\textbf{Why not Localize-MI.}
Localize-MI is sEEG with subject-specific intracranial electrode layouts. Subject-$i$'s 4D Gaussian field is fit using subject-$i$'s implant geometry (256 electrodes located inside that particular brain). Querying that field at subject-$j$'s electrode positions is only well defined if the two implants are co-registered to a common space, which the Localize-MI release does not provide consistently across all 7 subjects. We therefore restrict the cross-subject study to the two scalp-EEG datasets (SEED, SEED-IV) which share the same standard 62-channel cap.

\subsection{Cross-subject results}
\noindent\textbf{Full results.}
Table~\ref{tab:supp_crosssubj_full} reports mean$\pm$std across 3 split seeds for each (dataset, $r$). The variance is small (typically $<10^{-2}$ on PCC), so all conclusions in Section~\ref{sec:experiments:crosssubj} hold under any single seed.

\begin{table}[htbp]
  \caption{Full cross-subject results (mean$\pm$std over 3 split seeds).
    \emph{Within} is the diagonal of the $S\times S$ transfer matrix
    (matches Table~\ref{tab:eeg_results}).
    \emph{Cross} is the off-diagonal mean ($S(S{-}1)$ ordered pairs per
    seed). \emph{LD-spline} is non-learned spherical-spline upsampling of
    the LD montage. NMSE$\downarrow$, PCC$\uparrow$, SNR$\uparrow$ (dB).}
  \label{tab:supp_crosssubj_full}
  \centering
  \small
  \resizebox{0.75\columnwidth}{!}{%
  \begin{tabular}{ll l ccc}
    \toprule
    Dataset & $r$ & Metric & EMAG within & EMAG cross & LD-spline \\
    \midrule
    \multirow{9}{*}{SEED}
      & \multirow{3}{*}{$2\times$}
        & NMSE & $0.1545{\scriptstyle\pm0.0011}$ & $0.8203{\scriptstyle\pm0.0023}$ & $1.4895{\scriptstyle\pm0.0037}$ \\
      & & PCC  & $0.9194{\scriptstyle\pm0.0006}$ & $0.5392{\scriptstyle\pm0.0013}$ & $0.2552{\scriptstyle\pm0.0019}$ \\
      & & SNR  & $8.20{\scriptstyle\pm0.03}$ & $0.92{\scriptstyle\pm0.01}$ & $-1.72{\scriptstyle\pm0.01}$ \\
      \cmidrule(lr){2-6}
      & \multirow{3}{*}{$4\times$}
        & NMSE & $0.2776{\scriptstyle\pm0.0015}$ & $0.7819{\scriptstyle\pm0.0020}$ & $1.4447{\scriptstyle\pm0.0039}$ \\
      & & PCC  & $0.8496{\scriptstyle\pm0.0009}$ & $0.5349{\scriptstyle\pm0.0011}$ & $0.2776{\scriptstyle\pm0.0019}$ \\
      & & SNR  & $5.62{\scriptstyle\pm0.02}$ & $1.13{\scriptstyle\pm0.01}$ & $-1.58{\scriptstyle\pm0.01}$ \\
      \cmidrule(lr){2-6}
      & \multirow{3}{*}{$8\times$}
        & NMSE & $0.4215{\scriptstyle\pm0.0032}$ & $0.8613{\scriptstyle\pm0.0017}$ & $1.4371{\scriptstyle\pm0.0032}$ \\
      & & PCC  & $0.7595{\scriptstyle\pm0.0021}$ & $0.4603{\scriptstyle\pm0.0014}$ & $0.2815{\scriptstyle\pm0.0016}$ \\
      & & SNR  & $3.80{\scriptstyle\pm0.03}$ & $0.72{\scriptstyle\pm0.01}$ & $-1.56{\scriptstyle\pm0.01}$ \\
    \midrule
    \multirow{9}{*}{SEED-IV}
      & \multirow{3}{*}{$2\times$}
        & NMSE & $0.1151{\scriptstyle\pm0.0011}$ & $0.7848{\scriptstyle\pm0.0042}$ & $1.0625{\scriptstyle\pm0.0022}$ \\
      & & PCC  & $0.9405{\scriptstyle\pm0.0006}$ & $0.5651{\scriptstyle\pm0.0016}$ & $0.4687{\scriptstyle\pm0.0011}$ \\
      & & SNR  & $9.59{\scriptstyle\pm0.04}$ & $1.13{\scriptstyle\pm0.02}$ & $-0.24{\scriptstyle\pm0.01}$ \\
      \cmidrule(lr){2-6}
      & \multirow{3}{*}{$4\times$}
        & NMSE & $0.2023{\scriptstyle\pm0.0008}$ & $0.7840{\scriptstyle\pm0.0025}$ & $1.1350{\scriptstyle\pm0.0015}$ \\
      & & PCC  & $0.8928{\scriptstyle\pm0.0004}$ & $0.5493{\scriptstyle\pm0.0004}$ & $0.4325{\scriptstyle\pm0.0007}$ \\
      & & SNR  & $7.07{\scriptstyle\pm0.02}$ & $1.13{\scriptstyle\pm0.01}$ & $-0.52{\scriptstyle\pm0.01}$ \\
      \cmidrule(lr){2-6}
      & \multirow{3}{*}{$8\times$}
        & NMSE & $0.3358{\scriptstyle\pm0.0011}$ & $0.8519{\scriptstyle\pm0.0011}$ & $1.2668{\scriptstyle\pm0.0011}$ \\
      & & PCC  & $0.8143{\scriptstyle\pm0.0007}$ & $0.4870{\scriptstyle\pm0.0007}$ & $0.3666{\scriptstyle\pm0.0005}$ \\
      & & SNR  & $4.80{\scriptstyle\pm0.01}$ & $0.79{\scriptstyle\pm0.01}$ & $-0.94{\scriptstyle\pm0.01}$ \\
    \bottomrule
  \end{tabular}}
\end{table}

\noindent\textbf{Practical implication.}
EMAG should be deployed as a personalized SR module: a small per-subject fit (${\sim}76$k parameters, single-GPU minutes per subject) is required to unlock the high-fidelity within-subject regime, but the same architecture already provides a non-trivial uplift over classical interpolation in a zero-shot transfer setting, and could serve as a strong initialization for fast few-shot personalization---an avenue we leave to future work.

\begin{figure}[htbp]
  \centering
  \includegraphics[width=0.75\columnwidth]{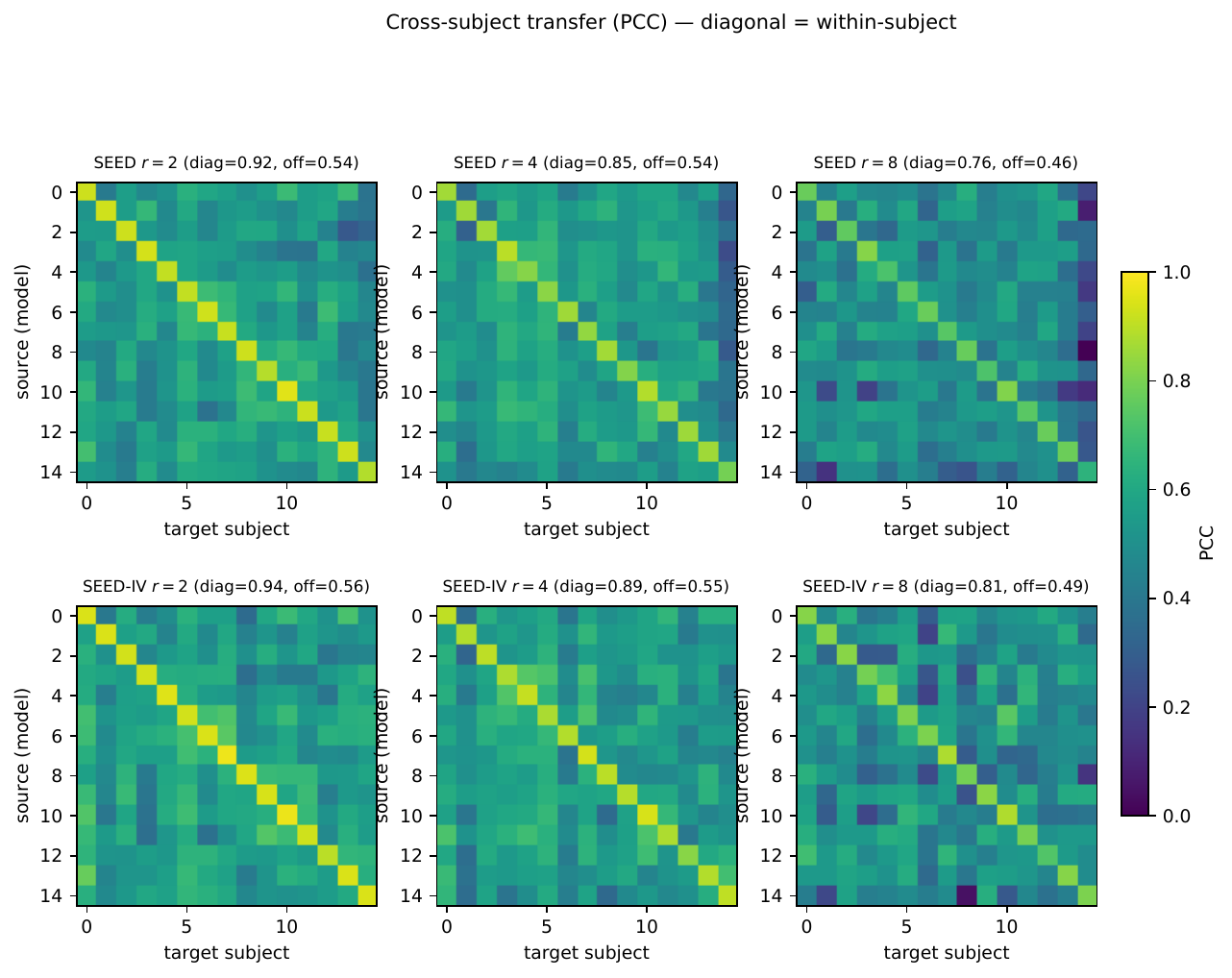}
  \caption{Subject$\times$subject PCC transfer matrices. Rows: source
    (training) subject; columns: target (evaluation) subject. Diagonal =
    within-subject score; off-diagonal mean = cross-subject score
    reported in Table~\ref{tab:supp_crosssubj_full}.}
  \label{fig:crosssubj_heatmaps}
\end{figure}

\noindent\textbf{Cross-subject transfer matrices} Figure~\ref{fig:crosssubj_heatmaps} shows the full $15{\times}15$ subject$\times$subject PCC transfer matrices used to derive the within/cross/LD-spline summary in Table~\ref{tab:supp_crosssubj_full}. The bright diagonal vs.\ pale off-diagonal pattern is consistent across both datasets and all SR factors, reaffirming that EMAG's per-subject fit captures features that do not transfer.

\noindent\textbf{Interpretation.}
Three observations are worth emphasizing. \textit{(i) Cross-subject EMAG strictly dominates LD-spline on all tested (dataset, $r$, metric) configurations.} The improvement is largest on SEED (PCC $0.46$\,--\,$0.54$ vs.\ $0.26$\,--\,$0.28$; NMSE $\sim\!0.78$\,--\,$0.86$ vs.\ $\sim\!1.44$\,--\,$1.49$) where the LD-spline baseline is weak, and remains clear on SEED-IV. This says that the physics-style structure baked into EMAG---a sphere-bounded mixture of anisotropic 4D Gaussians whose centers act as virtual cortical sources, projected to scalp through subject electrode geometry---generalizes across people without any retraining. \textit{(ii) The within-vs-cross gap is large} (e.g.\ on SEED-IV $r{=}2$, PCC $0.94 \to 0.57$, SNR $9.6\to 1.1$\,dB). Per-subject training thus captures, beyond the geometric prior, individual features of the source configuration (cortical folding, electrode-to-cortex distances, time-locked dipolar dynamics) that cannot be reused across subjects. This matches the design intent: EMAG's component centers are \emph{not} shared across subjects. \textit{(iii) The cross-subject score is nearly flat in $r$} (SEED-IV PCC $0.565{\to}0.549{\to}0.487$; SEED $0.539{\to}0.535{\to}0.460$), whereas the within-subject score degrades sharply with $r$. This is consistent with cross-subject performance being limited by thestructural mismatch between two brains rather than by the LD information content---the latter only becomes the bottleneck once subject identity is matched.

\subsection{Per-subject reconstruction quality}
\label{app:crosssubj:persubject}
Tables~\ref{tab:persubject_localize_mi}\,--\,\ref{tab:persubject_seed_iv}
report per-subject NMSE/PCC/SNR for every SR factor, using the default
$R{=}12$, $G{=}3$ sphere-grid configuration at split seed $0$.
Per-subject variability is moderate on Localize-MI (e.g.\ at $r{=}4$,
NMSE ranges from $0.07$ for sub-02 to $0.18$ for sub-04, the latter being a
known outlier with denser implant coverage in motor cortex), and tighter on
SEED/SEED-IV where the shared 62-channel cap normalizes input geometry.
Reconstruction quality degrades smoothly with $r$: on Localize-MI the mean
PCC drops $0.956 \to 0.946 \to 0.927 \to 0.885$ from $r{=}2$ to $r{=}16$.

\begin{table}[htbp]
\caption{Per-subject EMAG reconstruction on \textbf{Localize-MI} (split seed 0, $R{=}12$, $G{=}3$, sphere grid). NMSE $\downarrow$, PCC $\uparrow$, SNR $\uparrow$ (dB).}
\label{tab:persubject_localize_mi}
\centering\small
\resizebox{\columnwidth}{!}{
\begin{tabular}{l ccc ccc ccc ccc}
\toprule
Subject & \multicolumn{3}{c}{$2\times$} & \multicolumn{3}{c}{$4\times$} & \multicolumn{3}{c}{$8\times$} & \multicolumn{3}{c}{$16\times$} \\
\cmidrule(lr){2-4} \cmidrule(lr){5-7} \cmidrule(lr){8-10} \cmidrule(lr){11-13}
 & NMSE & PCC & SNR & NMSE & PCC & SNR & NMSE & PCC & SNR & NMSE & PCC & SNR \\
\midrule
1 & 0.068 & 0.965 & 11.67 & 0.086 & 0.956 & 10.64 & 0.110 & 0.943 & 9.59 & 0.161 & 0.916 & 7.93 \\
2 & 0.053 & 0.973 & 12.75 & 0.070 & 0.964 & 11.53 & 0.107 & 0.945 & 9.69 & 0.179 & 0.906 & 7.47 \\
3 & 0.070 & 0.964 & 11.52 & 0.086 & 0.956 & 10.65 & 0.126 & 0.935 & 8.99 & 0.177 & 0.907 & 7.52 \\
4 & 0.154 & 0.920 & 8.13 & 0.182 & 0.905 & 7.41 & 0.222 & 0.882 & 6.54 & 0.316 & 0.827 & 5.00 \\
5 & 0.066 & 0.966 & 11.80 & 0.081 & 0.959 & 10.93 & 0.114 & 0.941 & 9.44 & 0.190 & 0.900 & 7.20 \\
6 & 0.097 & 0.950 & 10.12 & 0.114 & 0.941 & 9.43 & 0.150 & 0.922 & 8.24 & 0.253 & 0.864 & 5.97 \\
7 & 0.091 & 0.953 & 10.39 & 0.108 & 0.945 & 9.68 & 0.156 & 0.919 & 8.08 & 0.229 & 0.878 & 6.40 \\
\midrule \textbf{Mean} & \textbf{0.086} & \textbf{0.956} & \textbf{10.91} & \textbf{0.104} & \textbf{0.946} & \textbf{10.04} & \textbf{0.141} & \textbf{0.927} & \textbf{8.65} & \textbf{0.215} & \textbf{0.885} & \textbf{6.78} \\
\bottomrule
\end{tabular}}
\end{table}

\begin{table}[htbp]
\caption{Per-subject EMAG reconstruction on \textbf{SEED} (split seed 0, $R{=}12$, $G{=}3$, sphere grid). NMSE $\downarrow$, PCC $\uparrow$, SNR $\uparrow$ (dB).}
\label{tab:persubject_seed}
\centering\small
\resizebox{0.75\columnwidth}{!}{
\begin{tabular}{l ccc ccc ccc}
\toprule
Subject & \multicolumn{3}{c}{$2\times$} & \multicolumn{3}{c}{$4\times$} & \multicolumn{3}{c}{$8\times$} \\
\cmidrule(lr){2-4} \cmidrule(lr){5-7} \cmidrule(lr){8-10}
 & NMSE & PCC & SNR & NMSE & PCC & SNR & NMSE & PCC & SNR \\
\midrule
1 & 0.148 & 0.923 & 8.29 & 0.256 & 0.863 & 5.92 & 0.403 & 0.773 & 3.95 \\
2 & 0.142 & 0.926 & 8.49 & 0.262 & 0.859 & 5.81 & 0.365 & 0.797 & 4.38 \\
3 & 0.143 & 0.926 & 8.43 & 0.257 & 0.862 & 5.91 & 0.428 & 0.756 & 3.69 \\
4 & 0.121 & 0.937 & 9.16 & 0.193 & 0.898 & 7.14 & 0.333 & 0.817 & 4.77 \\
5 & 0.173 & 0.910 & 7.63 & 0.332 & 0.817 & 4.79 & 0.485 & 0.718 & 3.14 \\
6 & 0.177 & 0.907 & 7.53 & 0.319 & 0.825 & 4.96 & 0.434 & 0.752 & 3.63 \\
7 & 0.136 & 0.929 & 8.65 & 0.262 & 0.859 & 5.82 & 0.405 & 0.771 & 3.92 \\
8 & 0.163 & 0.915 & 7.88 & 0.300 & 0.837 & 5.23 & 0.453 & 0.740 & 3.44 \\
9 & 0.144 & 0.925 & 8.41 & 0.260 & 0.860 & 5.85 & 0.410 & 0.768 & 3.88 \\
10 & 0.209 & 0.889 & 6.80 & 0.337 & 0.814 & 4.72 & 0.483 & 0.719 & 3.16 \\
11 & 0.088 & 0.955 & 10.55 & 0.214 & 0.887 & 6.70 & 0.345 & 0.810 & 4.63 \\
12 & 0.157 & 0.918 & 8.05 & 0.285 & 0.846 & 5.46 & 0.445 & 0.745 & 3.52 \\
13 & 0.164 & 0.914 & 7.86 & 0.274 & 0.852 & 5.62 & 0.406 & 0.771 & 3.92 \\
14 & 0.149 & 0.923 & 8.27 & 0.267 & 0.856 & 5.74 & 0.396 & 0.777 & 4.02 \\
15 & 0.220 & 0.883 & 6.58 & 0.375 & 0.790 & 4.26 & 0.580 & 0.648 & 2.36 \\
\midrule \textbf{Mean} & \textbf{0.156} & \textbf{0.919} & \textbf{8.17} & \textbf{0.280} & \textbf{0.848} & \textbf{5.59} & \textbf{0.425} & \textbf{0.757} & \textbf{3.76} \\
\bottomrule
\end{tabular}}
\end{table}

\begin{table}[htbp]
\caption{Per-subject EMAG reconstruction on \textbf{SEED-IV} (split seed 0, $R{=}12$, $G{=}3$, sphere grid). NMSE $\downarrow$, PCC $\uparrow$, SNR $\uparrow$ (dB).}
\label{tab:persubject_seed_iv}
\centering\small
\resizebox{0.75\columnwidth}{!}{
\begin{tabular}{l ccc ccc ccc}
\toprule
Subject & \multicolumn{3}{c}{$2\times$} & \multicolumn{3}{c}{$4\times$} & \multicolumn{3}{c}{$8\times$} \\
\cmidrule(lr){2-4} \cmidrule(lr){5-7} \cmidrule(lr){8-10}
 & NMSE & PCC & SNR & NMSE & PCC & SNR & NMSE & PCC & SNR \\
\midrule
1 & 0.110 & 0.943 & 9.57 & 0.183 & 0.904 & 7.36 & 0.333 & 0.817 & 4.78 \\
2 & 0.111 & 0.943 & 9.56 & 0.217 & 0.885 & 6.64 & 0.324 & 0.822 & 4.90 \\
3 & 0.133 & 0.931 & 8.77 & 0.192 & 0.899 & 7.17 & 0.323 & 0.823 & 4.91 \\
4 & 0.136 & 0.930 & 8.67 & 0.231 & 0.877 & 6.37 & 0.373 & 0.792 & 4.28 \\
5 & 0.096 & 0.951 & 10.16 & 0.168 & 0.912 & 7.74 & 0.314 & 0.828 & 5.03 \\
6 & 0.130 & 0.933 & 8.85 & 0.241 & 0.871 & 6.18 & 0.352 & 0.805 & 4.53 \\
7 & 0.113 & 0.942 & 9.45 & 0.216 & 0.885 & 6.66 & 0.353 & 0.804 & 4.52 \\
8 & 0.048 & 0.976 & 13.15 & 0.127 & 0.934 & 8.95 & 0.228 & 0.879 & 6.42 \\
9 & 0.103 & 0.947 & 9.85 & 0.195 & 0.897 & 7.10 & 0.374 & 0.791 & 4.27 \\
10 & 0.120 & 0.938 & 9.21 & 0.208 & 0.890 & 6.81 & 0.319 & 0.825 & 4.96 \\
11 & 0.061 & 0.969 & 12.15 & 0.117 & 0.940 & 9.32 & 0.225 & 0.880 & 6.47 \\
12 & 0.138 & 0.928 & 8.60 & 0.231 & 0.877 & 6.36 & 0.366 & 0.796 & 4.36 \\
13 & 0.196 & 0.897 & 7.07 & 0.324 & 0.822 & 4.89 & 0.470 & 0.728 & 3.28 \\
14 & 0.107 & 0.945 & 9.71 & 0.217 & 0.885 & 6.64 & 0.328 & 0.820 & 4.84 \\
15 & 0.112 & 0.942 & 9.50 & 0.182 & 0.905 & 7.41 & 0.357 & 0.802 & 4.47 \\
\midrule \textbf{Mean} & \textbf{0.114} & \textbf{0.941} & \textbf{9.62} & \textbf{0.203} & \textbf{0.892} & \textbf{7.04} & \textbf{0.336} & \textbf{0.814} & \textbf{4.80} \\
\bottomrule
\end{tabular}}
\end{table}

\section{Electrode-subset super-resolution: extended results}
\label{sec:elecsub:supp}

This appendix backs the electrode-subset study of Section~\ref{sec:elecsub}. We give (i) the prior-work motivation behind each named subset, (ii) the full channel lists, (iii) the experimental protocol, and (iv) per-(dataset, SR-factor, subset) fidelity numbers including SNR.

\subsection{Electrode geometry}
\label{app:qualitative_examples:electrodes}
\begin{figure}[htbp]
  \centering
  \includegraphics[width=\textwidth]{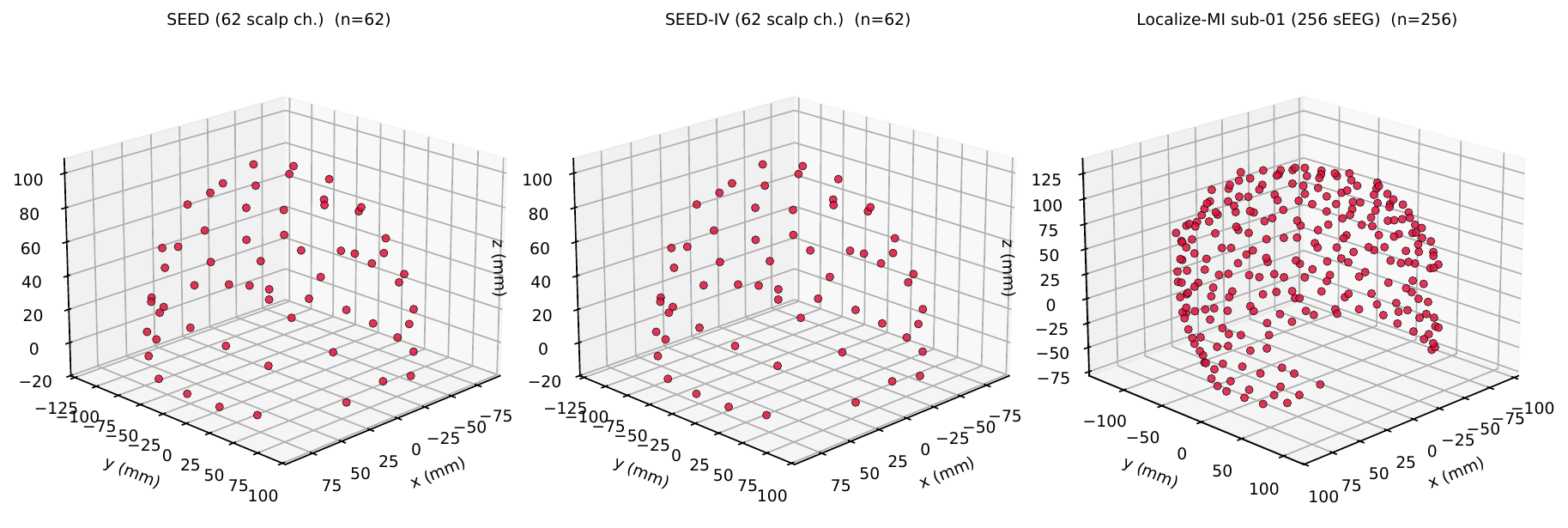}
  \caption{Recording geometries used in this paper. SEED/SEED-IV (left, middle) use the standard 10/20-extended 62-channel scalp cap; Localize-MI sub-01 (right) uses an intracerebral montage with $256$ contacts.}
  \label{fig:electrode_layouts}
\end{figure}

Figure~\ref{fig:electrode_layouts} shows the three recording geometries to scale: SEED and SEED-IV share a 62-channel scalp cap, while Localize-MI uses subject-specific intracranial implants ($256$ contacts for sub-01).
This visual contrast underlies many of the dataset-dependent patterns in the ablation study (Sec.~\ref{sec:ablation}), including the larger benefit of the volumetric brain grid on SEED.

\subsection{Background and motivation for the named subsets}
\label{sec:elecsub:app:motivation}

Channel selection has been an active subfield of EEG-based BCI design for over two decades; recent surveys review supervised, filter-, and wrapper-based methods that reduce the number of required channels without sacrificing classifier accuracy \citep{abdullah2022eeg,kim2022miniaturization}. For affective EEG specifically, two complementary priors recur in the literature.

\noindent\textbf{Frontal hemispheric asymmetry.}
\citet{davidson1992anterior} established the now-canonical link between asymmetric frontal alpha activity and approach / withdrawal motivation, which has motivated emotion-EEG datasets such as SEED~\citep{zheng2015investigating} to emphasise frontal and lateral coverage. Our \textsc{Hemi-Left} / \textsc{Hemi-Right} subsets at $K{=}31$ operationalise this prior by retaining one full hemisphere of the montage plus the four midline channels (Fz, Cz, Pz, Oz).

\noindent\textbf{Circumference / rim channels.}
\citet{valderrama2025identifying} argue that ten ``circumference'' channels of the 10/05 montage (Fp1, Fp2, F7, F8, T7, T8, P7, P8, O1, O2) carry most of the information used by direct emotion classifiers, and report that classifiers trained on these ten channels match the accuracy of classifiers trained on the full 62-channel montage on SEED. We use this finding as the inspiration for our \textsc{V15} subset (the ten circumference channels plus five additional rim electrodes, yielding $K{=}15$) and for the four hemispheric rim slices at $K{=}7$ (\textsc{VL7}, \textsc{VR7}, \textsc{VU7}, \textsc{VLw7}). The \textsc{FT15} subset is a fronto-temporal variant that emphasises channels typically associated with affective processing.

\noindent\textbf{Interior controls.}
For each rim subset we include a same-$K$ interior control (\textsc{INT15}, \textsc{INT7}) consisting of central / midline electrodes. These controls let us isolate the effect of \emph{rim versus interior} placement at fixed budget.

\subsection{Subset definitions}
\label{sec:elecsub:app:defs}
Each named subset is a list of channel labels from the canonical 62-channel SEED / SEED-IV ordering. Subset sizes are $K\in\{31, 15, 7\}$, giving SR factors of $\{2, 4, 8\}$ relative to the full montage.

\begin{table}[htbp]
  \centering
  \footnotesize
  \begin{tabular}{l c p{0.62\linewidth}}
    \toprule
    Subset ID    & $K$ & Channels (SEED order) \\
    \midrule
    \textsc{Hemi-Left}  & 31 & FP1, AF3, F7, F5, F3, F1, FT7, FC5, FC3, FC1, T7, C5, C3, C1, TP7, CP5, CP3, CP1, P7, P5, P3, P1, PO7, PO5, PO3, CB1, O1, FZ, CZ, PZ, OZ \\
    \textsc{Hemi-Right} & 31 & FP2, AF4, F8, F6, F4, F2, FT8, FC6, FC4, FC2, T8, C6, C4, C2, TP8, CP6, CP4, CP2, P8, P6, P4, P2, PO8, PO6, PO4, CB2, O2, FZ, CZ, PZ, OZ \\
    \textsc{V15}        & 15 & FP1, FP2, F7, F8, FT7, FT8, T7, T8, TP7, TP8, P7, P8, PO7, O1, O2 \\
    \textsc{FT15}       & 15 & FP1, FP2, F3, F4, F7, F8, FC5, FC6, FT7, FT8, T7, T8, TP7, TP8, P7 \\
    \textsc{INT15}      & 15 & AF3, AF4, F1, F2, FZ, FC1, FC2, C1, C2, CZ, CP1, CP2, P1, P2, PZ \\
    \textsc{VL7}        & 7  & FP1, F7, FT7, T7, TP7, P7, O1 \\
    \textsc{VR7}        & 7  & FP2, F8, FT8, T8, TP8, P8, O2 \\
    \textsc{VU7}        & 7  & FP1, FP2, F7, F8, FT7, FT8, T7 \\
    \textsc{VLw7}       & 7  & TP7, TP8, P7, P8, PO7, O1, O2 \\
    \textsc{INT7}       & 7  & FZ, FC1, FC2, CZ, CP1, CP2, PZ \\
    \bottomrule
  \end{tabular}
  \caption{Definitions of the named electrode subsets. \textsc{V15} is a
  superset of the ten circumference channels of
  \citet{valderrama2025identifying}. \textsc{Hemi-Left} and
  \textsc{Hemi-Right} retain the four midline channels (Fz, Cz, Pz, Oz) in both subsets to keep the size at exactly $K{=}31$.}
  \label{tab:elecsub:defs}
\end{table}

\begin{figure}[htbp]
  \centering
  \includegraphics[width=\columnwidth]{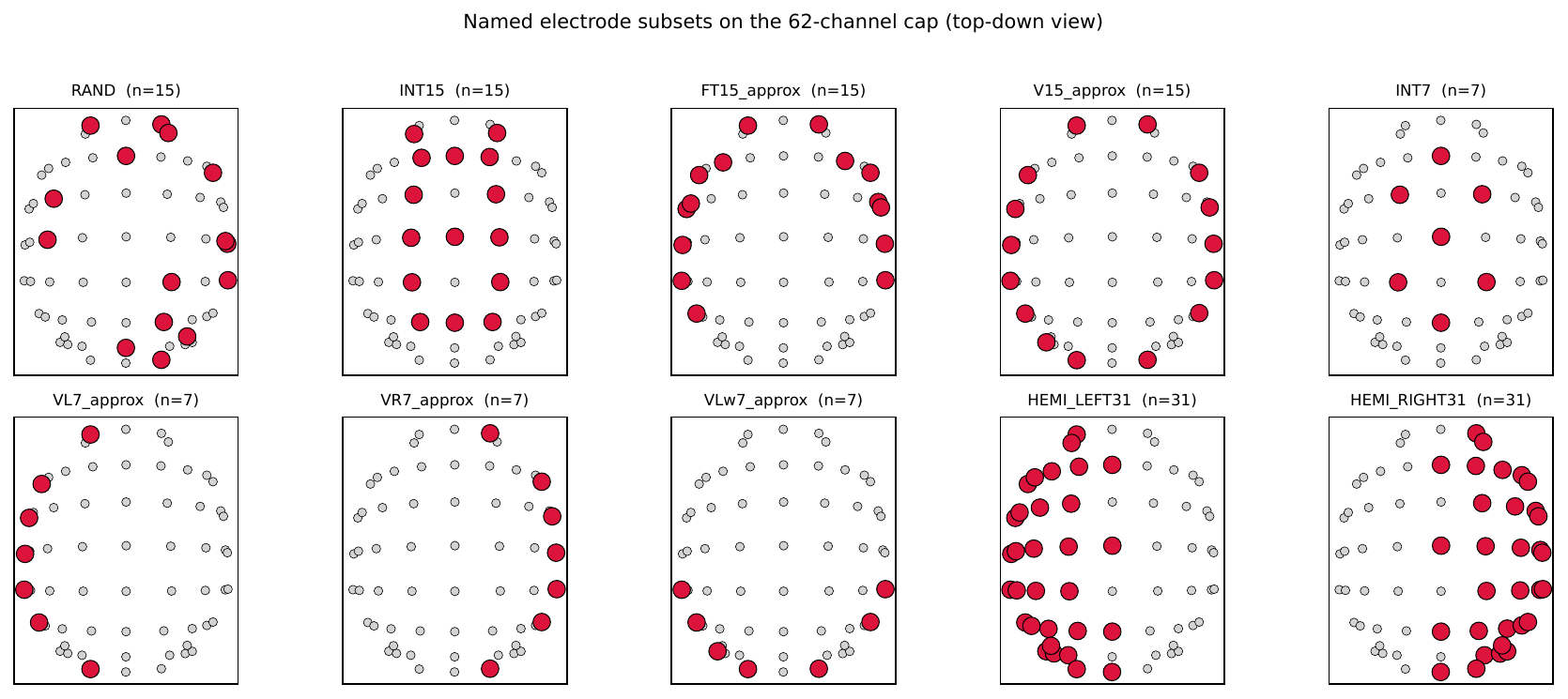}
  \caption{Named electrode subsets on the 62-channel SEED/SEED-IV cap
    (top-down view, anterior up). Crimson = retained channels, gray =
    discarded.}
  \label{fig:subset_coverage}
\end{figure}

\subsection{Protocol}
\label{sec:elecsub:app:protocol}

For every (dataset, SR-factor, subset) cell we trained EMAG with the standard configuration of the main paper: mixture of $G{=}3$ anisotropic 4D Gaussians per active grid point, $12^3$ uniform brain grid, full-4D covariance, leadfield-matrix forward operator, batch size 1, 100 epochs with patience 20 on validation NMSE. The \emph{only} quantity that varies across cells is the set of retained low-density channels: for the named subsets, this set is fixed to the channel list of Table~\ref{tab:elecsub:defs}; for the \textsc{Random} baseline, the $K$-subset is sampled uniformly at random per (subject, split seed). Every cell is trained from scratch -- the \textsc{Random} baseline does \emph{not} share weights with the main-benchmark random-subset model -- so the Random-versus-named comparison is a clean test of channel selection isolated from any compute-budget confound.

We repeat each (dataset, SR-factor, subset, subject) configuration across three split seeds $s\in\{0,1,2\}$, which control both the train/val/test split and -- for the \textsc{Random} baseline -- the random $K$-subset. The numbers reported in Section~\ref{sec:elecsub} and in Table~\ref{tab:elecsub:fidelity_full} below are the mean over the 15 subjects of the dataset, with mean$\pm$std taken across the three split seeds.

\subsection{Full reconstruction-fidelity table}
\label{sec:elecsub:app:table}

Table~\ref{tab:elecsub:fidelity_full} reports NMSE, PCC, and SNR for every (dataset, SR, subset) cell as mean$\pm$std over the three split seeds. The $\Delta_{\text{rand}}$ columns are absolute changes relative to the \textsc{Random} baseline at the same (dataset, SR) cell; positive $\Delta_{\text{rand}}\text{NMSE}$ and negative $\Delta_{\text{rand}}\text{PCC}$ / $\Delta_{\text{rand}}\text{SNR}$ all indicate that the named subset is worse than \textsc{Random}.

\begin{table}[htbp]
  \centering
  \scriptsize
  \setlength{\tabcolsep}{3pt}
  \resizebox{\textwidth}{!}{%
  \begin{tabular}{l l l ccc ccc}
    \toprule
    & & & \multicolumn{3}{c}{Metric (mean$\pm$std over 3 seeds)}
        & \multicolumn{3}{c}{$\Delta_{\text{rand}}$ (vs.\ \textsc{Random}, same SR)} \\
    \cmidrule(lr){4-6}\cmidrule(lr){7-9}
    Dataset & SR & Subset
      & NMSE $\downarrow$ & PCC $\uparrow$ & SNR (dB) $\uparrow$
      & $\Delta$NMSE & $\Delta$PCC & $\Delta$SNR \\
    \midrule
    SEED & $\times 2$ & \textsc{Random}
      & $0.154{\scriptstyle\pm0.001}$ & $0.919{\scriptstyle\pm0.001}$ & $8.20{\scriptstyle\pm0.03}$
      & $0.000$ & $0.000$ & $0.00$ \\
    SEED & $\times 2$ & \textsc{Hemi-Left}
      & $0.312{\scriptstyle\pm0.058}$ & $0.829{\scriptstyle\pm0.035}$ & $5.15{\scriptstyle\pm0.77}$
      & $+0.158$ & $-0.090$ & $-3.05$ \\
    SEED & $\times 2$ & \textsc{Hemi-Right}
      & $0.312{\scriptstyle\pm0.050}$ & $0.829{\scriptstyle\pm0.031}$ & $5.13{\scriptstyle\pm0.68}$
      & $+0.158$ & $-0.091$ & $-3.08$ \\
    \midrule
    SEED & $\times 4$ & \textsc{Random}
      & $0.278{\scriptstyle\pm0.002}$ & $0.850{\scriptstyle\pm0.001}$ & $5.62{\scriptstyle\pm0.02}$
      & $0.000$ & $0.000$ & $0.00$ \\
    SEED & $\times 4$ & \textsc{V15}
      & $0.384{\scriptstyle\pm0.037}$ & $0.784{\scriptstyle\pm0.024}$ & $4.22{\scriptstyle\pm0.42}$
      & $+0.106$ & $-0.066$ & $-1.41$ \\
    SEED & $\times 4$ & \textsc{FT15}
      & $0.385{\scriptstyle\pm0.038}$ & $0.783{\scriptstyle\pm0.024}$ & $4.19{\scriptstyle\pm0.42}$
      & $+0.108$ & $-0.066$ & $-1.44$ \\
    SEED & $\times 4$ & \textsc{INT15}
      & $0.418{\scriptstyle\pm0.037}$ & $0.762{\scriptstyle\pm0.024}$ & $3.83{\scriptstyle\pm0.38}$
      & $+0.140$ & $-0.087$ & $-1.79$ \\
    \midrule
    SEED & $\times 8$ & \textsc{Random}
      & $0.422{\scriptstyle\pm0.003}$ & $0.760{\scriptstyle\pm0.002}$ & $3.80{\scriptstyle\pm0.03}$
      & $0.000$ & $0.000$ & $0.00$ \\
    SEED & $\times 8$ & \textsc{INT7}
      & $0.529{\scriptstyle\pm0.006}$ & $0.685{\scriptstyle\pm0.005}$ & $2.78{\scriptstyle\pm0.05}$
      & $+0.108$ & $-0.074$ & $-1.01$ \\
    SEED & $\times 8$ & \textsc{VR7}
      & $0.527{\scriptstyle\pm0.001}$ & $0.687{\scriptstyle\pm0.001}$ & $2.80{\scriptstyle\pm0.01}$
      & $+0.105$ & $-0.072$ & $-0.99$ \\
    SEED & $\times 8$ & \textsc{VL7}
      & $0.540{\scriptstyle\pm0.017}$ & $0.677{\scriptstyle\pm0.013}$ & $2.70{\scriptstyle\pm0.14}$
      & $+0.118$ & $-0.082$ & $-1.09$ \\
    SEED & $\times 8$ & \textsc{VU7}
      & $0.561{\scriptstyle\pm0.002}$ & $0.662{\scriptstyle\pm0.002}$ & $2.53{\scriptstyle\pm0.02}$
      & $+0.139$ & $-0.098$ & $-1.26$ \\
    SEED & $\times 8$ & \textsc{VLw7}
      & $0.572{\scriptstyle\pm0.004}$ & $0.654{\scriptstyle\pm0.003}$ & $2.44{\scriptstyle\pm0.03}$
      & $+0.150$ & $-0.106$ & $-1.35$ \\
    \midrule
    SEED-IV & $\times 2$ & \textsc{Random}
      & $0.115{\scriptstyle\pm0.001}$ & $0.940{\scriptstyle\pm0.001}$ & $9.58{\scriptstyle\pm0.03}$
      & $0.000$ & $0.000$ & $0.00$ \\
    SEED-IV & $\times 2$ & \textsc{Hemi-Right}
      & $0.275{\scriptstyle\pm0.032}$ & $0.851{\scriptstyle\pm0.019}$ & $5.67{\scriptstyle\pm0.50}$
      & $+0.160$ & $-0.090$ & $-3.91$ \\
    SEED-IV & $\times 2$ & \textsc{Hemi-Left}
      & $0.285{\scriptstyle\pm0.030}$ & $0.845{\scriptstyle\pm0.018}$ & $5.50{\scriptstyle\pm0.45}$
      & $+0.170$ & $-0.095$ & $-4.08$ \\
    \midrule
    SEED-IV & $\times 4$ & \textsc{Random}
      & $0.203{\scriptstyle\pm0.001}$ & $0.893{\scriptstyle\pm0.000}$ & $7.06{\scriptstyle\pm0.02}$
      & $0.000$ & $0.000$ & $0.00$ \\
    SEED-IV & $\times 4$ & \textsc{V15}
      & $0.299{\scriptstyle\pm0.010}$ & $0.837{\scriptstyle\pm0.006}$ & $5.29{\scriptstyle\pm0.15}$
      & $+0.097$ & $-0.056$ & $-1.77$ \\
    SEED-IV & $\times 4$ & \textsc{FT15}
      & $0.311{\scriptstyle\pm0.020}$ & $0.829{\scriptstyle\pm0.012}$ & $5.12{\scriptstyle\pm0.28}$
      & $+0.108$ & $-0.063$ & $-1.94$ \\
    SEED-IV & $\times 4$ & \textsc{INT15}
      & $0.330{\scriptstyle\pm0.025}$ & $0.818{\scriptstyle\pm0.015}$ & $4.89{\scriptstyle\pm0.33}$
      & $+0.127$ & $-0.075$ & $-2.17$ \\
    \midrule
    SEED-IV & $\times 8$ & \textsc{Random}
      & $0.336{\scriptstyle\pm0.001}$ & $0.814{\scriptstyle\pm0.001}$ & $4.80{\scriptstyle\pm0.01}$
      & $0.000$ & $0.000$ & $0.00$ \\
    SEED-IV & $\times 8$ & \textsc{INT7}
      & $0.441{\scriptstyle\pm0.001}$ & $0.747{\scriptstyle\pm0.001}$ & $3.61{\scriptstyle\pm0.01}$
      & $+0.105$ & $-0.067$ & $-1.19$ \\
    SEED-IV & $\times 8$ & \textsc{VLw7}
      & $0.469{\scriptstyle\pm0.005}$ & $0.727{\scriptstyle\pm0.003}$ & $3.34{\scriptstyle\pm0.05}$
      & $+0.134$ & $-0.087$ & $-1.46$ \\
    SEED-IV & $\times 8$ & \textsc{VU7}
      & $0.470{\scriptstyle\pm0.001}$ & $0.727{\scriptstyle\pm0.001}$ & $3.32{\scriptstyle\pm0.01}$
      & $+0.135$ & $-0.087$ & $-1.48$ \\
    SEED-IV & $\times 8$ & \textsc{VR7}
      & $0.491{\scriptstyle\pm0.000}$ & $0.713{\scriptstyle\pm0.000}$ & $3.10{\scriptstyle\pm0.00}$
      & $+0.156$ & $-0.102$ & $-1.70$ \\
    SEED-IV & $\times 8$ & \textsc{VL7}
      & $0.501{\scriptstyle\pm0.001}$ & $0.706{\scriptstyle\pm0.001}$ & $3.02{\scriptstyle\pm0.01}$
      & $+0.165$ & $-0.108$ & $-1.78$ \\
    \bottomrule
  \end{tabular}}
  \caption{Three-seed reconstruction-fidelity numbers for the electrode-subset experiments. Each cell is the mean over the 15 subjects of the dataset, reported as mean$\pm$std across split seeds $s\in\{0,1,2\}$. The \textsc{Random} rows are the same uniform-random baseline employed throughout the paper, retrained at each split seed with an independently re-sampled $K$-subset per (subject, seed).}
  \label{tab:elecsub:fidelity_full}
\end{table}

\subsection{Discussion}
\label{sec:elecsub:app:discussion}

\noindent\textbf{Why does random uniform sampling beat anatomical priors?}
The named subsets are designed for direct decoding: they concentrate electrodes on regions whose univariate or low-order statistics are discriminative for the downstream task. The SR objective is different; it asks the reconstructor to recover the entire scalp signal, including regions that the named subsets leave entirely unobserved. Under EMAG's forward model, each retained electrode constrains the implicit source field within roughly a Gaussian-shaped neighbourhood centred on the electrode position; if a large region of the scalp has no nearby observed channel, the reconstructor must extrapolate and pays an NMSE / PCC penalty. Uniform random sampling spreads observations isotropically across the scalp, which is close to the optimal coverage pattern under this model.

\noindent\textbf{Why does the rim-vs-interior gap shrink at SR$\times$8?}
At $K{=}7$ the absolute number of observed channels is small enough that any clustered subset -- rim or interior -- leaves the rest of the scalp under-sampled. The total spatial coverage of \textsc{INT7} (which
spreads its seven electrodes along the central midline strip) turns out to be comparable to that of the rim slices at this regime, which is consistent with the small NMSE differences we observe in
Table~\ref{tab:elecsub:fidelity_full}.

\section{Ablations: extended results and discussion}
\label{app:ablations}
This section gives the full ablation results and per-component discussion underlying Section~\ref{sec:ablation}.
We conduct a systematic ablation study to validate each core design  decision in EMAG. All ablations are evaluated on the Localize-MI and SEED  datasets using the $4\times$ super-resolution factor, as this setting provides a representative balance between reconstruction difficulty and spatial ambiguity. We report NMSE, PCC, and SNR on the held-out test set, averaged across all subjects. Results are summarized in the Table~\ref{tab:combined_ablation}.

\subsection{Hyperparameter search}
\label{app:hpsearch}
We searched grid resolution $R\!\in\!\{5,7,10,12,15\}$ and Gaussians-per-point $G\!\in\!\{1,2,3,5,8\}$ on Localize-MI sub-01 at $r{=}4$ with a TPE search (\textsc{Optuna}); learning rate was jointly sampled in $\{1{\times}10^{-4},\ldots,5{\times}10^{-3}\}$. For every $(R,G)$ pair we report the lowest validation NMSE achieved across the sampled learning rates.

\begin{table}[h]
\centering
\small
\caption{Hyperparameter search on Localize-MI ($r{=}4$): validation NMSE
($\downarrow$) by grid resolution $R$ (rows) and Gaussians-per-point $G$
(columns). Bold = chosen configuration.}
\label{tab:hp:RxG}
\begin{tabular}{c|ccccc}
\toprule
$R\backslash G$ & 1 & 2 & 3 & 5 & 8 \\ \midrule
5  & $0.150$ & $0.164$ & $0.244$ & $0.405$ & $0.405$ \\
7  & $0.142$ & $0.143$ & $0.132$ & $0.274$ & $0.136$ \\
10 & $0.131$ & $0.131$ & $0.115$ & $0.128$ & $0.136$ \\
12 & $0.123$ & $0.114$ & $\mathbf{0.106}$ & $0.131$ & $0.115$ \\
15 & $0.113$ & $0.113$ & $0.113$ & $0.131$ & $0.107$ \\
\bottomrule
\end{tabular}
\end{table}

Performance saturates by $R{=}12$, with diminishing returns at $R{=}15$ (and $\sim2{\times}$ compute). Increasing $G$ beyond 3 does not improve fidelity at $R{\geq}10$. Learning-rate sensitivity was narrow: all top-quartile trials used $\mathrm{lr}\!\in\![5{\times}10^{-4}, 3{\times}10^{-3}]$; we adopt $\mathrm{lr}{=}1{\times}10^{-3}$ throughout. Final headline configuration: $R{=}12,\,G{=}3$ (\textbf{bold}).

\subsection{4D Gaussian Parameterization}
\label{sec:ablation:parameterization}

The distinguishing structural element of EMAG is its use of full $4\times4$ anisotropic precision matrices for each Gaussian component. We isolate the contribution of each aspect of this parameterization through three targeted ablations.

\noindent\textbf{Temporal dimension in the precision matrix.}
We first ask whether the fourth (temporal) axis of the precision matrix carries information beyond what amplitude modulation alone provides. To test this, we replace the full $4\times4$ precision matrix with a $3\times3$ spatial-only block, removing the temporal dimension from the Mahalanobis distance computation in Eq.~\ref{eq:forward} entirely. In this variant, temporal dynamics are handled exclusively by the time-varying amplitude $a_n(t)$.

We find that removing the temporal axis consistently degrades performance across both datasets. The larger relative degradation on SEED is consistent with the richer spectral content of continuous resting-state recordings, where transient activations with complex spatiotemporal profiles are more common than in the short, stereotyped evoked responses of Localize-MI. In both cases the degradation confirms that the temporal covariance axis encodes information that amplitude modulation alone cannot recover.

\noindent\textbf{Off-diagonal coupling terms.}
We next isolate the contribution of anisotropy and space-time coupling by zeroing out all off-diagonal entries $c_0, \ldots, c_5$ in the Cholesky factor $\mathbf{L}_n$ (Eq.~\ref{eq:cholesky}), reducing each component to an axis-aligned ellipsoid. We further decompose this into 
two sub-ablations: (a)~removing only the spatial off-diagonal terms $c_0, c_1, c_2$, preserving patial isotropy but retaining space-time coupling; and (b)~removing space-time terms $c_3,c_4,c_5$ while introducing spatial anisotropy.

Results across both datasets reveal a consistent pattern: removing \emph{any} subset of off-diagonal terms produces similar degradation, and the two sub-ablations are produce comparable lesser results. Two conclusions follow. First, both coupling types are \emph{jointly} necessary: either type alone recovers no advantage over the fully uncoupled case, indicating that their contributions are complementary rather than additive. Second, the full $4\times4$ precision matrix,  which enables all six coupling terms simultaneously, is strictly required to capture the interaction between source directionality and temporal evolution, and no lower-dimensional subset achieves an equivalent representation.

\noindent\textbf{Isotropic scalar variance.}
As the minimal parameterization within the Gaussian family, we replace each component's full precision matrix with a single learnable scalar width $\sigma_n$, yielding a fully isotropic Gaussian with one variance parameter per component. This corresponds to the isotropic variant described in Section~\ref{sec:experiments:results} and establishes the performance floor for the Gaussian family.

The isotropic variant produces the largest degradation within the parameterization category. Taken together with the off-diagonal blations, the results establish a clear ordering: full $4\times4$ $>$ diagonal anisotropic $>$ isotropic, with each step yielding a consistent performance improvement. The steeper decline on SEED suggests that the anisotropic parameterization is particularly important when source configurations are spatially distributed and heterogeneous, as in resting-state EEG, compared to the more focal, stereotyped activations of the stimulation paradigm in Localize-MI.

\subsection{Brain Grid and Source Placement}
\label{sec:ablation:grid}

EMAG anchors its Gaussian components on an anatomically-constrained 
spherical brain grid. We evaluate the necessity of this structural prior 
and its performance with other grid variations.

\noindent\textbf{Anatomical grid constraint.}
We remove the spherical constraint of Eq.~\ref{eq:braingrid} and allow all Gaussian centers to be free-floating learnable parameters, initialized at the original grid positions but unconstrained during training. Without the anatomical prior, both reconstruction quality and training stability degrade. Inspecting learned positions, a subset of free-floating components migrates outside the brain volume entirely, attaching to the scalp surface. These surface-hugging components act as shortcut interpolators, exploiting proximity to LD electrodes rather than learning meaningful source configurations. This phenomenon confirms that the brain grid constraint is not merely a computational convenience but a necessary regularizer that prevents the model from collapsing to a degenerate surface interpolation solution.

\noindent\textbf{Cortical surface vs. volumetric grid.}
To test whether deep (subcortical) sources contribute to reconstruction accuracy, we restrict grid points to the brain surface by retaining only those grid points within a thin shell of the outer sphere boundary. This surface-only model closely approximates cortical surface source models used in classical EEG analysis.

The impact of restricting sources to the brain surface is markedly dataset-dependent and considerably stronger than anticipated. On Localize-MI, performance degrades measurably, which is expected with the dataset's known subcortical stimulation sites that the surface-only model cannot represent. On SEED, however, the degradation is even greater, a finding which challenges the standard assumption that emotion-related EEG activity is predominantly cortical: the volumetric subcortical grid appears to be critical for effective SEED reconstruction. This ablation establishes that the full volumetric grid is the appropriate default and that surface-constrained source models are not a viable alternative for general-purpose EEG super-resolution.

\subsection{Low-Density Conditioning Architecture}
\label{sec:ablation:conditioning}

The LD conditioning pipeline is responsible for translating sparse 
observed electrode signals into spatially-resolved source amplitude 
modulations. We verify each component of this pipeline.

\noindent\textbf{No conditioning (static source template).}
We first establish the most critical baseline: removing the TemporalEncoder and MLP$_\text{mod}$ entirely, leaving each Gaussian with only its fixed base amplitude $w_{i,g}$ and no dependence on the LD input. In this degenerate setting, the model becomes a fixed anatomical template with learned spatial geometry but no ability to respond to the observed signal.

As expected, performance collapses under this model. This result confirms the self-evident but important fact that input-dependent conditioning is the primary driver of per-sample reconstruction accuracy, and that the Gaussian source geometry alone, while beneficial for generalization and interpretability, cannot substitute for dynamic signal-driven modulation. This ablation also serves as an interpretable lower bound: any method that approaches this performance level is effectively failing to use the LD input.

\noindent\textbf{Global vs.\ spatially-resolved amplitude modulation.}
We replace the per-grid-point amplitude output $\Delta w_i(t) \in 
\mathbb{R}^N$ (Eq.~\ref{eq:modulation}) with a single global scalar 
$\Delta w(t) \in \mathbb{R}$ applied uniformly to all grid points. This 
preserves the temporal conditioning mechanism but collapses spatial 
resolution: the model knows \emph{when} the brain is active but not 
\emph{where}.

The contrast between datasets here is the most striking in the entire ablation suite. On Localize-MI, collapsing the per-grid-point modulation to a global scalar causes a near-catastrophic degradation, while on SEED the same modification have a lesser effect. This dissociation directly reflects the nature of the two tasks. Localize-MI recordings consist of focal, spatially-specific evoked responses to single-site intracranial stimulation. SEED recordings, however, capture distributed, slower-varying emotion-related activity across the entire cortex. This result highlights an important design consideration: the benefit of spatially-resolved conditioning scales with the spatial heterogeneity and focal specificity of the neural signals, and is most critical in clinical or evoked-response settings.

\noindent\textbf{Raw vs.\ pre-interpolated LD input.}
We replace the raw $m$-channel LD input to the TemporalEncoder with a spherical-spline-interpolated version upsampled to the full $M$ HD electrode positions~\citep{perrin1989spherical}. This tests whether the encoder is bottlenecked by the sparsity of its input, and whether classical interpolation as a pre-processing step could substitute for the implicit spatial regularization provided by the Gaussian forward model.

We found that pre-interpolation yields no improvement over raw LD input. We attribute this result to the Gaussian source model acting as a strong spatial prior: because the forward model constrains electrode signals to arise from smooth volumetric source distributions, the encoder receives implicit spatial regularization through the backward pass during training, making external pre-interpolation redundant. This finding has a practical implication: EMAG does not require any electrode-space pre-processing and can operate directly on raw LD recordings.

\subsection{Physics-Informed Forward Model}
\label{sec:ablation:forward}

The differentiable Gaussian rendering pipeline is the architectural element that most distinguishes EMAG from prior electrode-space super-resolution methods. We evaluate its necessity and examine alternatives.

\noindent\textbf{Direct electrode-space mapping.}
We replace the entire source field rendering pipeline with a direct mapping: the TemporalEncoder output $\mathbf{h}_t$ is passed to an MLP that directly predicts the $M$-dimensional HD signal at each timestep, with no intermediate source representation. Model capacity is approximately matched to EMAG by widening the MLP hidden dimension. This ablation represents the strongest possible ``physics-free'' baseline within our framework and directly tests the central premise of the paper.

The benefit of the physics-informed Gaussian pipeline is dataset-specific but always present. On localize-MI, the direct mapping degrades with a substantial gap that confirms the value of the source-space bottleneck for focal, evoked-response data. On SEED, the gap narrows, which is consistent with the relatively smooth and distributed nature of emotion-related EEG: when the target signal lacks sharp spatial gradients, a capacity-matched MLP could learn an adequate electrode-space mapping from data alone. The contrast across datasets underscores that the physics-informed design is most valuable precisely in the settings where it is clinically important: when neural activity is spatially focal and the super-resolution fidelity is critical.

\noindent\textbf{Learned linear projection vs.\ Gaussian kernel.}
We replace the Gaussian-weighted distance function in Eq.~\ref{eq:forward} with a learned linear matrix $\mathbf{W} \in \mathbb{R}^{M \times N_{\text{total}}}$, trained end-to-end alongside the rest of the model. This preserves the source-space bottleneck representation but replaces the physically-grounded Gaussian kernel with an unconstrained linear projection, testing whether the specific \emph{form} of the Gaussian kernel is necessary or whether any differentiable source-to-electrode mapping suffices.

The learned linear projection behaves very differently across the two datasets, revealing an important generalization asymmetry. On Localize-MI, its scores nearly match those from the direct electrode mapping, landing far below EMAG. This indicates that the source-space bottleneck alone does not confer a meaningful advantage when the projection kernel is unconstrained. On SEED, the learned projection performs dramatically worse, and we hypothesize this stems from over-parameterization: the $M \times N_{\text{total}}$ weight matrix has no geometric prior and is free to memorize subject-specific electrode relationships. With 15 subjects and longer continuous recordings, the SEED training set exposes a wider variety of spatial configurations. Thus, the unconstrained projection fails to generalize across this diversity, whereas EMAG's Gaussian kernel imposes a distance-decay inductive bias that is invariant to specific electrode layouts and subjects. Together, the two datasets demonstrate that the \emph{form} of the Gaussian kernel, and not merely the existence of a source-space intermediate, greatly contributes to EMAG's generalization capabilities.

\noindent\textbf{Classical leadfield forward model.}
Finally, we replace the learned Gaussian kernel with a fixed electromagnetic leadfield matrix $\mathbf{L} \in \mathbb{R}^{M \times N_{\text{total}}}$ derived from a standard three-shell spherical head model using MNE-Python~\citep{gramfort2014mne}, keeping all other components of EMAG unchanged. The leadfield is computed at the same grid point locations as EMAG's brain grid, roviding a direct comparison between physics-derived and learned projection kernels.

The fixed leadfield model represents the lowest-performing forward model variant on both datasets. We argue that the performance gap relative to EMAG arises from multiple compounding factors: (1) the spherical three-shell head model does not account for individual anatomical variation, (2) the fixed kernel cannot adapt to the data, and (3) the BEM-derived conductivity assumptions do not hold for all subjects or electrode montages. Notably, however, the leadfield model does not improve interpretability in the manner one might expect from a physically-derived kernel, since the reconstruction error itself becomes large enough to obscure meaningful source configurations. This result reinforces the core design choice of EMAG: a \emph{learned} Gaussian kernel that is geometrically structured but data-adaptive outperforms both the fully unconstrained and the fully fixed alternatives, occupying a productive middle ground between flexibility and physical inductive bias.

\subsection{Ablation Summary}
\label{sec:ablation:summary}

Table~\ref{tab:combined_ablation}
consolidate all ablation results. Three overarching findings emerge. First, the single most impactful component is spatially-resolved LD conditioning: removing it entirely drives NMSE to $\approx 1.0$ on both datasets, while collapsing it to a global scalar causes a $+434\%$ NMSE increase on Localize-MI, underscoring that dynamic, spatially-specific modulation is indispensable for focal neural activity. Second, the full $4\times4$ precision matrix contributes meaningfully beyond isotropic baselines: the complete parameterization reduces NMSE by 0.0192 on Localize-MI and 0.0612 on SEED relative to a scalar-width Gaussian, with both spatial anisotropy and space-time coupling terms jointly required. Third, the physics-informed Gaussian kernel's advantage over the direct mapping support our claim that embedding the EEG forward model as a differentiable rendering step is particularly beneficial for super-resolution, where data-driven interpolation is often constrained. Taken together, these results validate each major design decision in EMAG: anatomical grid anchoring, volumetric source placement, full 4D anisotropic covariance, and spatially-resolved conditioning are all necessary components, each contributing independently to the overall performance.

\section{Interpretability of learned Gaussians}
\label{app:interp}

\begin{figure}[htbp]
  \centering
  \includegraphics[width=0.85\columnwidth]{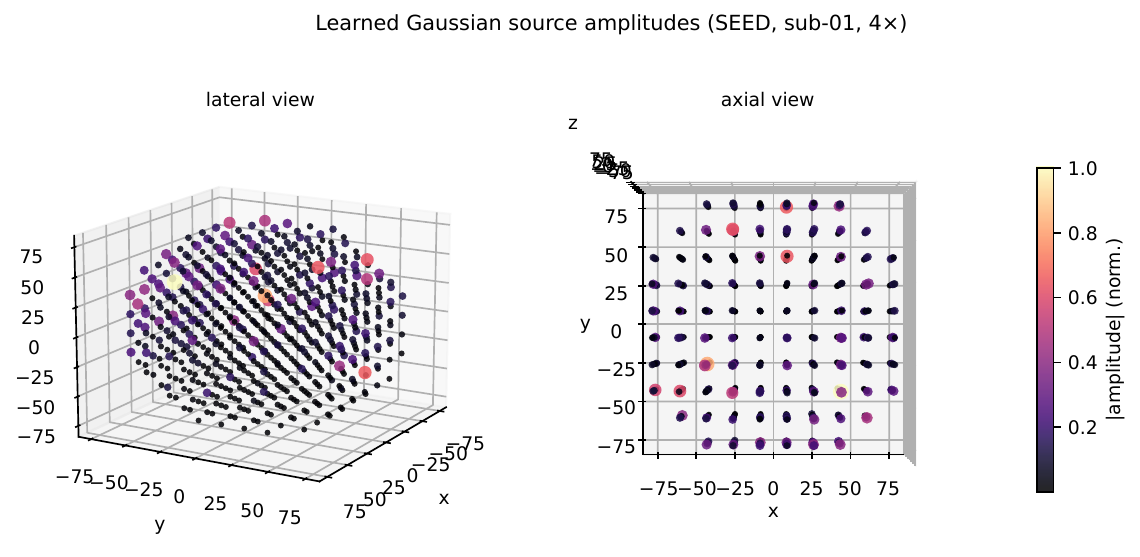}\\[2pt]
  \includegraphics[width=0.85\columnwidth]{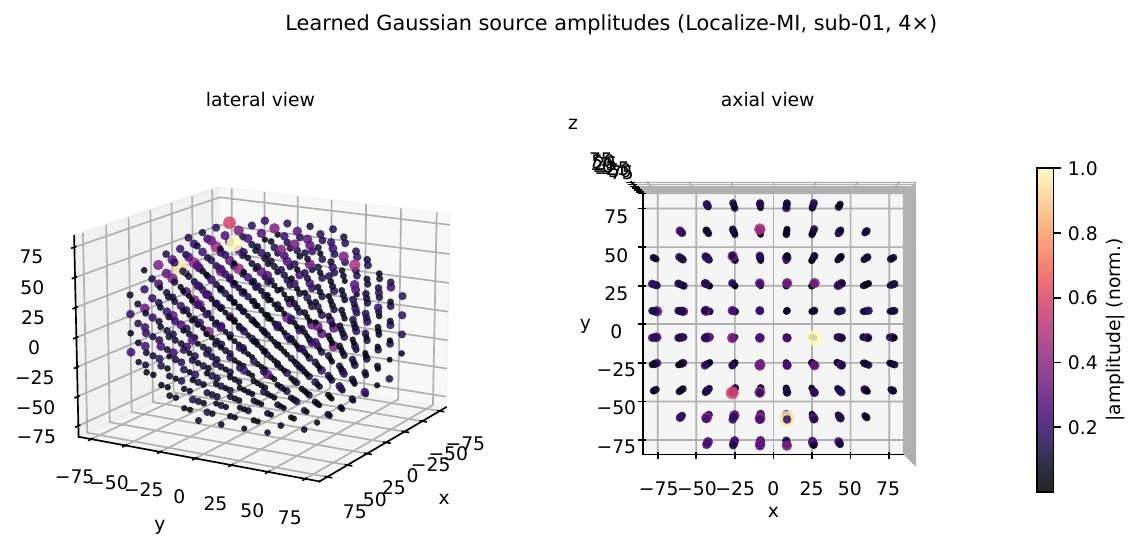}
  \caption{Learned Gaussian amplitudes (lateral and axial views) for SEED
    sub-01 (top) and Localize-MI sub-01 (bottom), trained at $r{=}4$ with
    the default configuration. Colour and size encode the mean amplitude
    magnitude across the three Gaussians anchored at each grid point.}
  \label{fig:gaussian_sources}
\end{figure}

The learned amplitudes of the $\sim\!2{,}500$ Gaussian components reveal where the model concentrates explanatory power inside the brain volume. Figure~\ref{fig:gaussian_sources} renders the active grid points of two trained subjects (one per dataset) coloured by mean amplitude magnitude across the $G{=}3$ Gaussians per point. On SEED (sub-01), high-amplitude components form a distributed pattern with clear bilateral occipital and frontal foci, matching the prior on emotion-related EEG. On Localize-MI (sub-01), the learned amplitudes concentrate near the intracerebral implant sites, consistent with the focal stimulation paradigm of the dataset.

\subsection{Learned Gaussian source maps}
\label{app:interp:sources}

\noindent\textbf{Quantitative validation against stimulation sites.}
We test whether the trained mixture allocates capacity near the ground-truth source. For each Localize-MI subject we load the trained
checkpoint ($r{=}4$, default config) and score every Gaussian by its LD-conditioned post-stim energy
\[
s_n \;=\; \mathbb{E}_{\text{trial},\,t\in[\,t_\mathrm{stim},\,t_\mathrm{end}\,]}
\!\big[(w_n+m_n(\mathrm{LD}(t)))^{2}\big],
\]
where $w_n$ is the base amplitude, $m_n(\cdot)$ the LD modulator output (Section~\ref{sec:method:ld}), and $t_\mathrm{stim}$ the stimulation onset of the run. For every run we then compute the Euclidean distance (mm, T1w-native space) from each top-$K$ ($K{=}10$) Gaussian centre to the active sEEG-pair midpoint $(\mathbf{p}_{e_1}+\mathbf{p}_{e_2})/2$. As a chance baseline we draw $5{,}000$ random points from the brain grid and report the mean distance and the empirical $p$-value $\Pr[d_\mathrm{rand}\le d_\mathrm{top1}]$. Per-run distances are averaged within subject; we then report the mean and standard deviation across the seven subject means.

\begin{figure}[htbp]
\centering
\includegraphics[width=\textwidth]{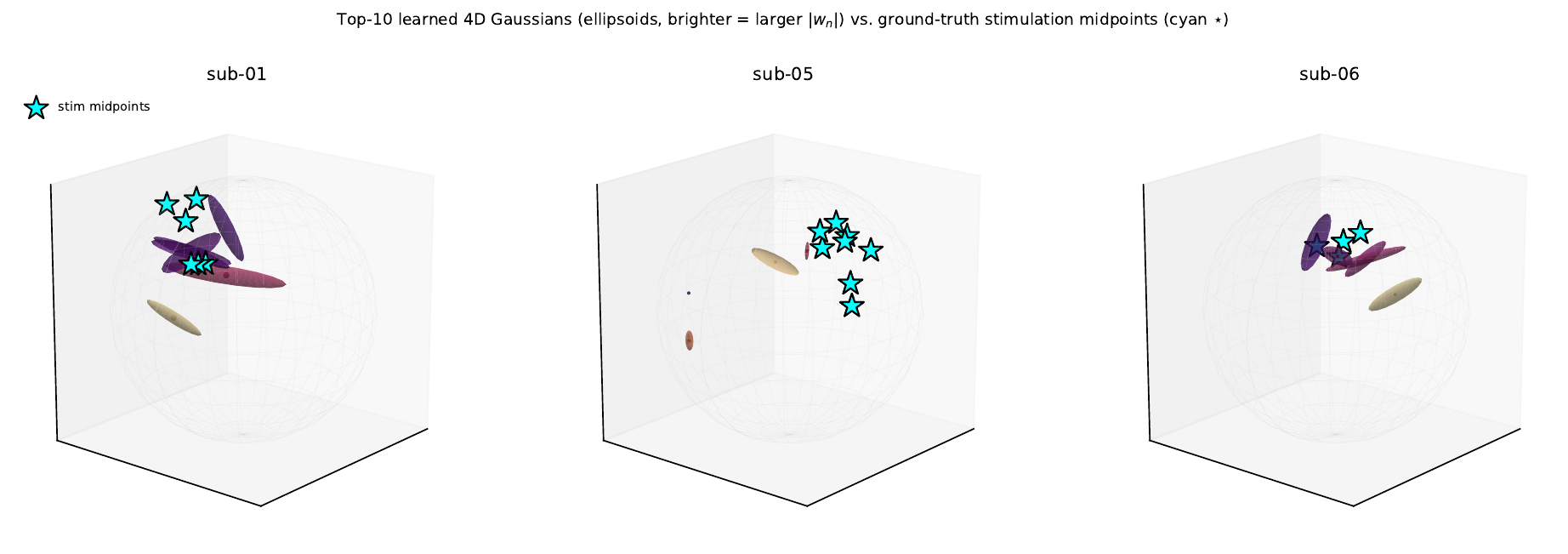}
\caption{Top-10 learned 4D Gaussians rendered as $1\sigma$ spatial ellipsoids (brightness $\propto |w_n|$) versus ground-truth stimulation midpoints (cyan $\star$), in T1w-native space inside the brain envelope ($r{=}90$\,mm). Subjects shown: best (sub-01), median (sub-05), and worst (sub-06).}
\label{fig:source_validation_3d}
\end{figure}

\begin{figure}[htbp]
\centering
\includegraphics[width=\textwidth]{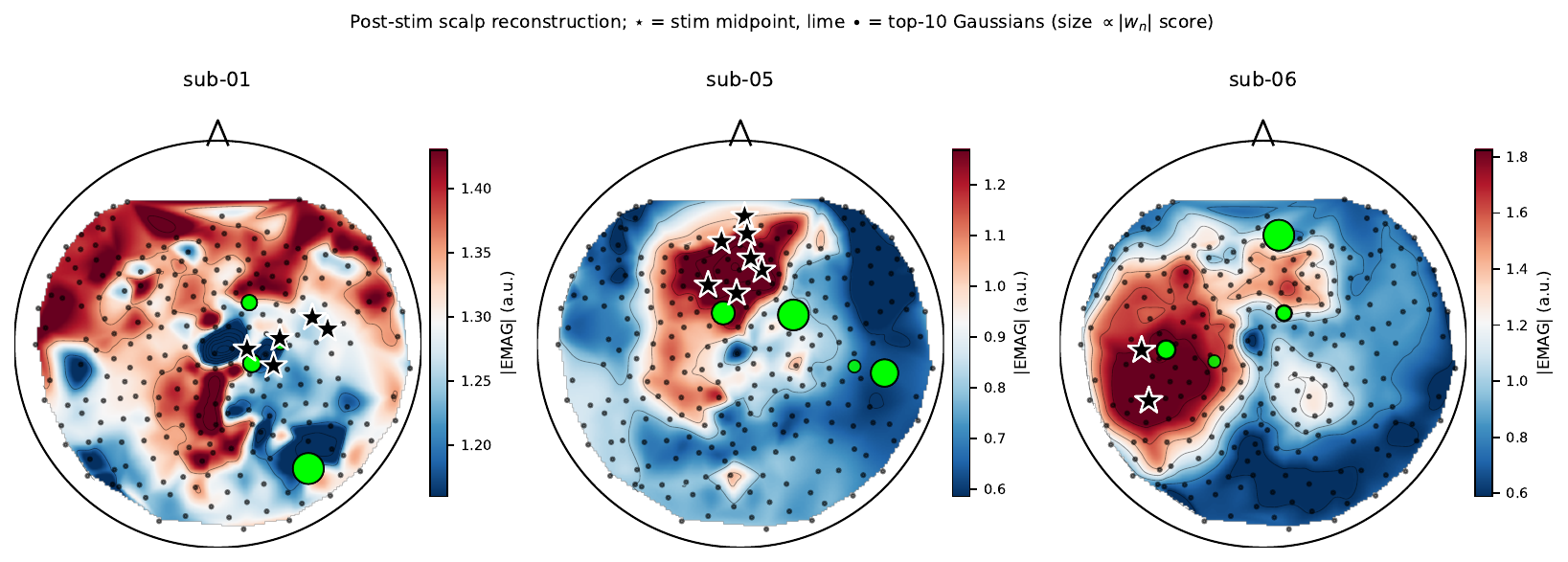}
\caption{Reconstructed scalp amplitude in the post-stim window (mean $|\mathrm{EMAG}|$ over $0.25$--$0.26$\,s, across test trials). Black $\star$ = scalp electrode nearest each run's stim midpoint; lime $\bullet$ = top-10 Gaussians projected azimuthally to the same layout, marker size $\propto$ post-stim LD-conditioned score. Reconstructed amplitude peaks consistently sit near both the stim-midpoint projection and the high-scoring Gaussian projections.}
\label{fig:source_validation_topomap}
\end{figure}

Across $7$ subjects and $61$ stim runs (Table~\ref{tab:source_val}), the nearest top-Gaussian sits at $d_1\!=\!42.1\!\pm\!14.4$\,mm from the stim midpoint, vs.\ a chance distance of $93.3\!\pm\!5.5$\,mm; the top-1 Gaussian lands in the best $10\!\pm\!8\%$ of the brain volume on average. The effect is consistent across subjects (per-subject $d_1$: $22$--$64$\,mm) and is achieved \emph{without any source-localization supervision} during training. A robustness check that ranks Gaussians by $|w_n|$ alone (no LD modulation) gives a near-identical headline ($d_1\!=\!42.3\!\pm\!11.9$ \,mm, $p\!=\!0.09\!\pm\!0.06$), confirming the finding does not depend on the conditioning term. Figure~\ref{fig:source_validation_3d} visualises the top-10 Gaussian ellipsoids against the stim midpoints, and Figure~\ref{fig:source_validation_topomap} shows the corresponding reconstructed scalp-amplitude topographies in the post-stim window with the stim-midpoint projection (black~$\star$) and the top-10 Gaussian projections (lime~$\bullet$, size~$\propto$~score) overlaid.

\begin{table}[htbp]
\centering
\small
\caption{Distance (mm) from learned top-$K$ Gaussians to ground-truth sEEG stimulation midpoints, Localize-MI ($r{=}4$, default config), $K{=}10$, ranked by post-stim LD-conditioned energy. $d_1$: nearest top-Gaussian; $\bar d_K$: mean over the top-$K$; $d_\mathrm{chance}$: mean distance from a random brain-grid point; $p$: empirical fraction of random points closer than $d_1$. Overall row = mean$\pm$std across the seven subject means.}
\label{tab:source_val}
\begin{tabular}{l|cccc}
\toprule
Subject & $d_1$ (mm) $\downarrow$ & $\bar d_K$ (mm) & $d_\mathrm{chance}$ (mm) & $p$ \\ \midrule
sub-01 & $\mathbf{22.0}$ & $53.0$ & $93.5$  & $\mathbf{0.02}$ \\
sub-02 & $44.1$ & $66.8$ & $101.8$ & $0.06$ \\
sub-03 & $64.5$ & $83.1$ & $90.1$  & $0.23$ \\
sub-04 & $54.9$ & $97.1$ & $89.9$  & $0.18$ \\
sub-05 & $38.7$ & $72.1$ & $89.4$  & $0.08$ \\
sub-06 & $29.4$ & $64.0$ & $88.2$  & $0.04$ \\
sub-07 & $40.9$ & $56.5$ & $100.2$ & $0.09$ \\ \midrule
\textbf{Overall} & $\mathbf{42.1\!\pm\!14.4}$ & $\mathbf{70.4\!\pm\!15.4}$ & $\mathbf{93.3\!\pm\!5.5}$ & $\mathbf{0.10\!\pm\!0.08}$ \\
\bottomrule
\end{tabular}
\end{table}

\subsection{Anisotropy distribution}
\label{app:interp:anisotropy}
\begin{figure}[htbp]
  \centering
  \includegraphics[width=0.75\columnwidth]{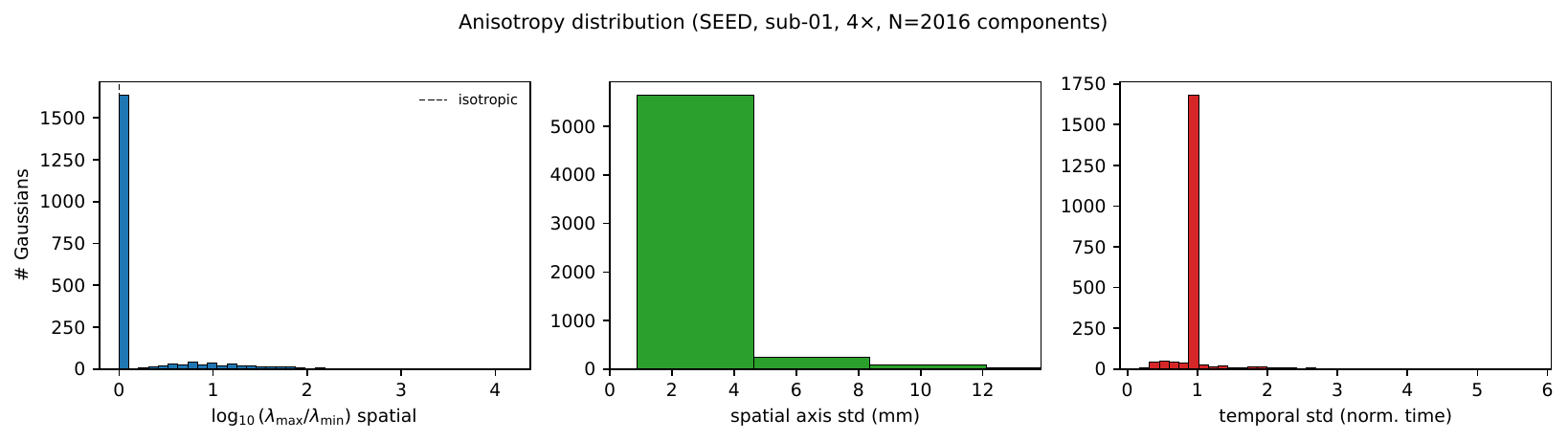}\\[2pt]
  \includegraphics[width=0.75\columnwidth]{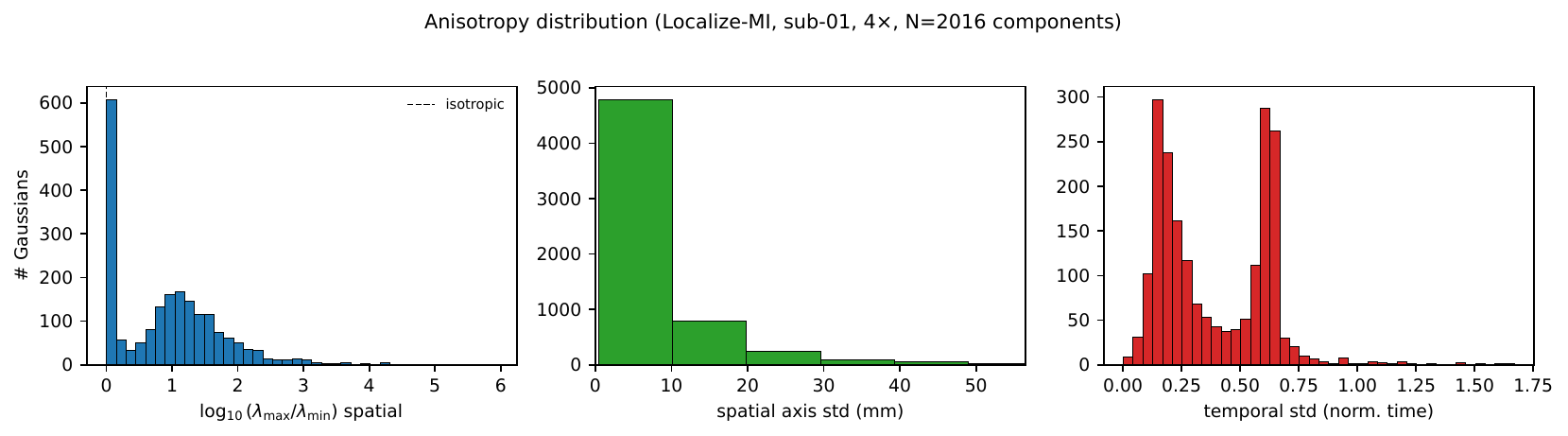}
  \caption{Anisotropy of the trained 4D Gaussian field (top: SEED sub-01,
    bottom: Localize-MI sub-01, both at $r{=}4$). Left: spatial
    log-condition-number $\log_{10}(\lambda_{\max}/\lambda_{\min})$ of the
    $3{\times}3$ spatial covariance block. Middle: per-axis spatial standard
    deviation in millimetres. Right: temporal standard deviation in
    normalized time units.}
  \label{fig:anisotropy}
\end{figure}

Figure~\ref{fig:anisotropy} shows distributions of (a)~the log-condition-number of the spatial $3{\times}3$ covariance block, (b)~the per-axis spatial standard deviation, and (c)~the temporal standard deviation of the $4{\times}4$ Cholesky-parameterized covariance. The anisotropy histogram is heavy-tailed and clearly separated from $0$ (isotropic), confirming that components do not collapse to spheres at training time and supporting the parameterization ablation in Sec.~\ref{sec:ablation:parameterization}. Spatial spreads concentrate around $5$\,--\,$15$\,mm, in line with the $15$\,mm initialization, while temporal spreads cover the full normalized range, indicating that the temporal axis of the precision matrix specializes per component---a behaviour the diagonal/spatial-only ablations cannot reproduce.

\section{Qualitative examples}
\label{app:qualitative_examples}
\subsection{Waveform reconstruction examples}
\label{app:qualitative_examples:waveforms}
\begin{figure}[htbp]
  \centering
  \includegraphics[width=0.85\columnwidth]{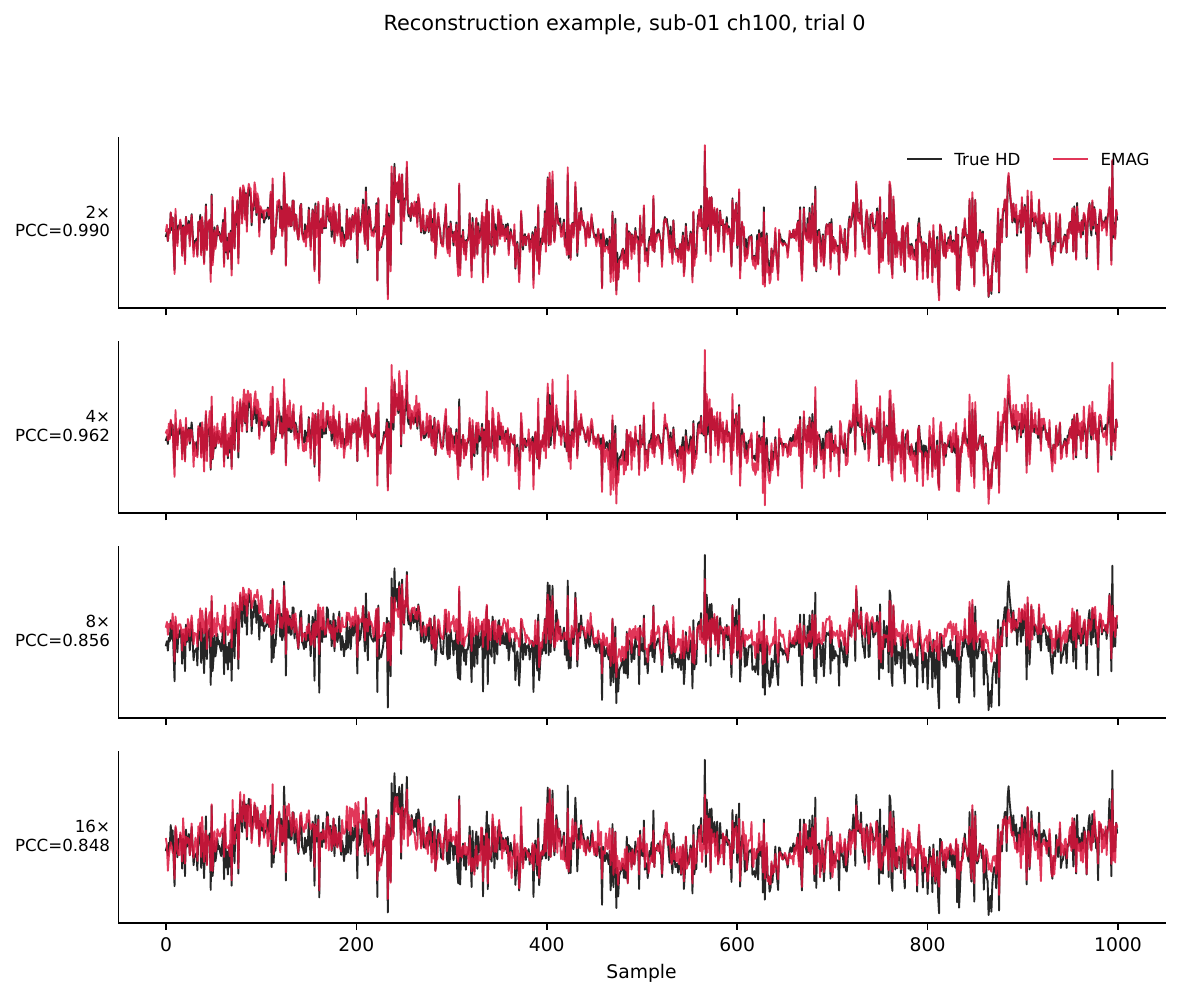}
  \caption{Localize-MI sub-01, channel 100, trial 0: ground-truth high-density
    signal (black) vs.\ EMAG reconstruction (crimson) at every SR factor.
    Per-trace PCC reported on the left.}
  \label{fig:reconstruction_example}
\end{figure}

Figure~\ref{fig:reconstruction_example} plots a representative high-density channel of sub-01 at every SR factor. EMAG tracks the slow envelope and the millisecond-scale stimulation artefacts at $r{=}2$\,--\,$8$, with visible amplitude attenuation at $r{=}16$ where only $16$ input channels remain.

\subsection{Spectral fidelity}
\label{app:qualitative_examples:psd}
\begin{figure}[htbp]
  \centering
  \includegraphics[width=0.75\columnwidth]{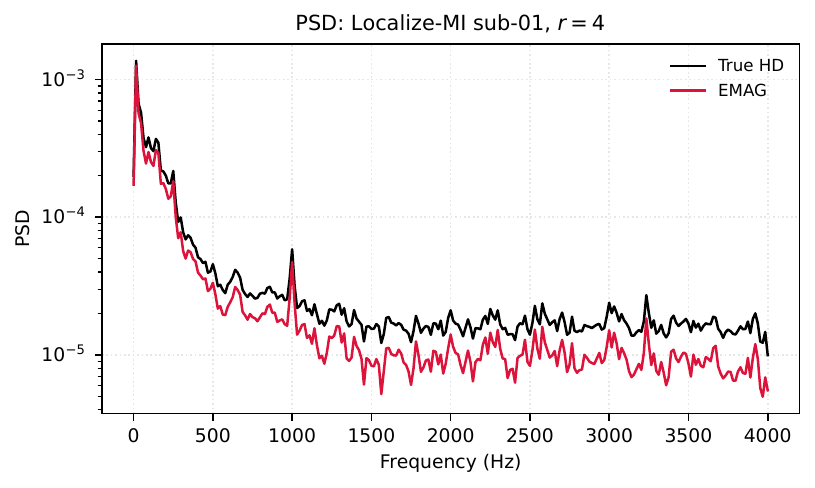}
  \includegraphics[width=0.75\columnwidth]{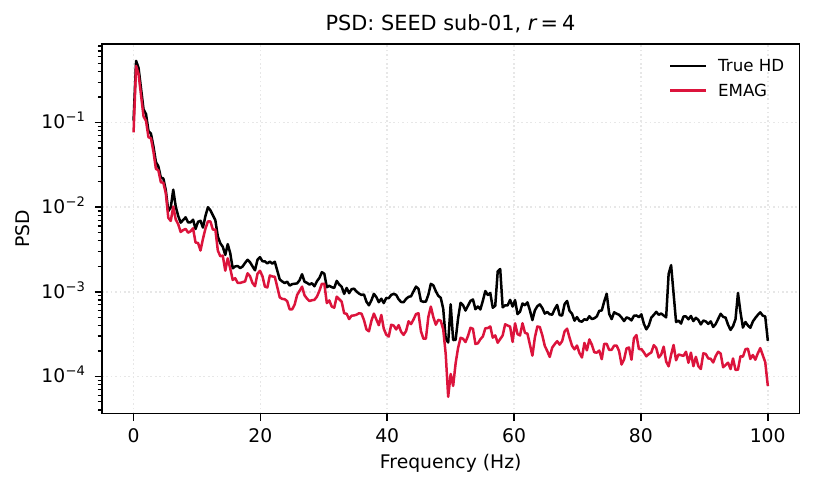}
  \caption{Welch PSD: true (black) vs.\ EMAG (crimson). Top:
    Localize-MI sub-01 ($f_s{=}8$\,kHz); bottom: SEED sub-01
    ($f_s{=}200$\,Hz). Both at $r{=}4$.}
  \label{fig:psd}
\end{figure}

Figure~\ref{fig:psd} compares Welch power-spectral densities of the true HD signal and EMAG's reconstruction, averaged over channels and trials. EMAG matches the broadband shape and reproduces the dataset-specific peaks (Localize-MI stimulation harmonics; SEED resting-state $\alpha$ band around $10$\,Hz), with mild high-frequency attenuation that is consistent with the smoothness prior of the Gaussian forward model.

\subsection{Single-time-step snapshot}
\label{app:qualitative_examples:snapshot}
\begin{figure}[htbp]
  \centering
  \includegraphics[width=0.75\columnwidth]{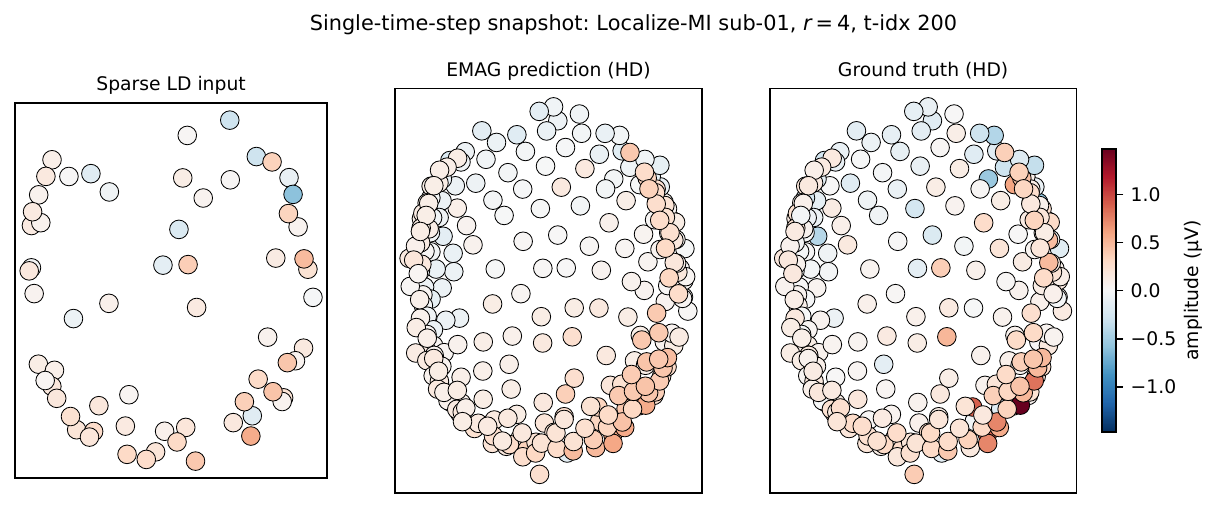}\\[2pt]
  \includegraphics[width=0.75\columnwidth]{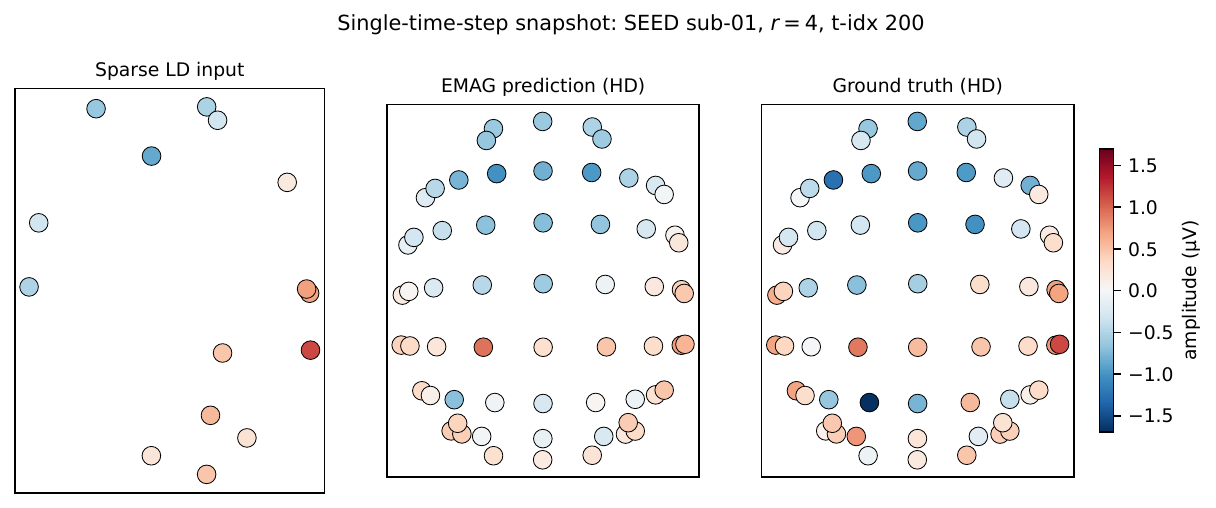}
  \caption{Single-time-step electrode-space snapshots
    (top: Localize-MI sub-01, bottom: SEED sub-01, $r{=}4$).
    Left: sparse LD input ($K{=}\lfloor n_{\rm hd}/r\rfloor$ retained
    channels). Middle: EMAG HD reconstruction. Right: ground truth.}
  \label{fig:snapshot}
\end{figure}

Figure~\ref{fig:snapshot} visualises one HD time step as a top-down electrode scatter, comparing the sparse LD input fed to EMAG, the EMAG HD reconstruction, and the ground-truth HD field. The reconstruction recovers spatial structure that is invisible in the LD input alone, including sign reversals across hemispheres on SEED. 

\section{Limitations and Impact}
\label{app:limitations}
\subsection{Limitations}
\label{app:limitations:limitations}
EMAG is trained per-subject and does not generalise zero-shot (Section~\ref{sec:experiments:crosssubj}). Evaluation is restricted to non-pathological scalp EEG; behaviour on clinical recordings, MEG, or intracranial signals is uncharacterised. Source-localisation fidelity is externally validated only on Localize-MI, where ground-truth stimulation sites are available. Compute scales with the number of active grid points; sub-second inference on edge hardware would require further optimisation.

\subsection{Broader impact}
\label{app:limitations:broader}
EMAG can lower the equipment cost of high-fidelity scalp EEG by recovering HD-quality recordings from sparse, consumer-grade caps, with positive implications for low-resource clinical and BCI use. The same capability introduces risks: reconstructions are model-based extrapolations, not direct measurements, and may create spurious confidence in clinical decisions taken on low-density recordings. We recommend downstream-task validation (e.g.\ classification accuracy, ERP-component fidelity) before any clinical deployment, and we explicitly discourage use of EMAG outputs as standalone diagnostic evidence.

\section{Ethics and Responsible Use}
\label{app:ethics}
This work uses only publicly released EEG datasets that were originally collected under the ethical approvals and licenses reported by their creators. We did not collect any new human-subject data, interact with participants, or perform any additional interventions beyond the use of these existing de-identified datasets. Because EEG signals are biometric and can potentially be sensitive, we treat reconstruction outputs as research artifacts rather than diagnostic evidence, and we caution that they should not be used for clinical decision-making without proper validation, regulatory review, and domain-specific oversight. More broadly, high-fidelity EEG super-resolution may improve low-cost neurotechnology and retrospective analysis, but it could also be misused to over-interpret sparse recordings or create unwarranted confidence in reconstructed signals. For this reason, we recommend that any downstream deployment clearly disclose the reconstruction process, report uncertainty where possible, and validate conclusions on the original observed data rather than on reconstructed signals alone.

\FloatBarrier
\newpage
\section*{NeurIPS Paper Checklist}

\begin{enumerate}

\item {\bf Claims}
    \item[] Question: Do the main claims made in the abstract and introduction accurately reflect the paper's contributions and scope?
    \item[] Answer: \answerYes{} 
    \item[] Justification: The abstract and Section~\ref{sec:intro} state EMAG's contributions (volumetric 4D-Gaussian EEG super-resolution on a spherical brain grid, anisotropic spatio-temporal covariance, LD-conditioned reconstruction); Section~\ref{sec:experiments:results} (Table~\ref{tab:eeg_results}) and Section~\ref{sec:elecsub} (Table~\ref{tab:elecsub:fidelity_main}) report the matching empirical claims; scope is restricted to the three well established datasets and SR factors evaluated.

\item {\bf Limitations}
    \item[] Question: Does the paper discuss the limitations of the work performed by the authors?
    \item[] Answer: \answerYes{} 
    \item[] Justification: Section~\ref{sec:discussion} discusses cross-subject generalization gap (Section~\ref{sec:experiments:crosssubj}), per-subject training requirement, restriction to non-pathological scalp EEG, and lack of source-localization ground truth outside Localize-MI.

\item {\bf Theory assumptions and proofs}
    \item[] Question: For each theoretical result, does the paper provide the full set of assumptions and a complete (and correct) proof?
    \item[] Answer: \answerNA{} 
    \item[] Justification: The paper presents an empirical method without theoretical results requiring proof; the rendering equations in Section~\ref{sec:method:mixture} follow standard Gaussian-mixture derivations (\citep{kerbl20233d}).

    \item {\bf Experimental result reproducibility}
    \item[] Question: Does the paper fully disclose all the information needed to reproduce the main experimental results of the paper to the extent that it affects the main claims and/or conclusions of the paper (regardless of whether the code and data are provided or not)?
    \item[] Answer: \answerYes{} 
    \item[] Justification: Architecture, loss, optimizer, schedule, grid resolution, Gaussians-per-point, and data splits are described in Section~\ref{sec:method} and App.~\ref{app:implementation}; full hyperparameter and ablation tables are in App.~\ref{app:ablations}; code and configs will be released at the public GitHub repository.

\item {\bf Open access to data and code}
    \item[] Question: Does the paper provide open access to the data and code, with sufficient instructions to faithfully reproduce the main experimental results, as described in supplemental material?
    \item[] Answer: \answerYes{} 
    \item[] Justification: Code, training scripts, and configs will be released at a public GitHub repository; the three datasets (Localize-MI, SEED, SEED-IV) are public and access procedures are described in App.~\ref{app:datasets}.


\item {\bf Experimental setting/details}
    \item[] Question: Does the paper specify all the training and test details (e.g., data splits, hyperparameters, how they were chosen, type of optimizer) necessary to understand the results?
    \item[] Answer: \answerYes{} 
    \item[] Justification: Section~\ref{sec:experiments:dataset} states splits, optimizer (Adam, \citep{kingma2014adam}), learning rate, batch size, and SR factors; full per-experiment hyperparameters are tabulated in App.~\ref{app:implementation}.

\item {\bf Experiment statistical significance}
    \item[] Question: Does the paper report error bars suitably and correctly defined or other appropriate information about the statistical significance of the experiments?
    \item[] Answer: \answerYes{} 
    \item[] Justification: All headline tables in Section~\ref{sec:experiments:results} (Table~\ref{tab:eeg_results}) and ablations (Table~\ref{tab:combined_ablation}) report mean$\pm$std over 3 seeds; per-subject standard deviations are tabulated in App.~\ref{app:crosssubj:persubject}.

\item {\bf Experiments compute resources}
    \item[] Question: For each experiment, does the paper provide sufficient information on the computer resources (type of compute workers, memory, time of execution) needed to reproduce the experiments?
    \item[] Answer: \answerYes{} 
    \item[] Justification: All experiments ran on a computing cluster (NVIDIA GPUs, 48GB VRAM class, e.g.\ RTX6000 nodes). Each (subject, SR-factor) EMAG run takes 30--120\,min wall-clock.
    
\item {\bf Code of ethics}
    \item[] Question: Does the research conducted in the paper conform, in every respect, with the NeurIPS Code of Ethics \url{https://neurips.cc/public/EthicsGuidelines}?
    \item[] Answer: \answerYes{} 
    \item[] Justification: The work uses only publicly released, de-identified EEG datasets under their original licenses; no new human-subjects data were collected and no foreseeable Code-of-Ethics violations apply.

\item {\bf Broader impacts}
    \item[] Question: Does the paper discuss both potential positive societal impacts and negative societal impacts of the work performed?
    \item[] Answer: \answerYes{} 
    \item[] Justification: Section~\ref{sec:discussion} discusses positive impacts (lower-cost BCI/clinical EEG via sparse caps, more accurate downstream interpretation) and risks (reconstructions could create spurious confidence in low-density recordings, potential misuse for pseudo-clinical decisions) and recommends downstream-task validation before deployment.
    
\item {\bf Safeguards}
    \item[] Question: Does the paper describe safeguards that have been put in place for responsible release of data or models that have a high risk for misuse (e.g., pre-trained language models, image generators, or scraped datasets)?
    \item[] Answer: \answerNA{} 
    \item[] Justification: The released model is a small ($<$100k-parameter) per-subject signal-reconstruction model trained on public EEG; it does not generate identifiable biometric content or pose dual-use concerns of the kind targeted by this question.

\item {\bf Licenses for existing assets}
    \item[] Question: Are the creators or original owners of assets (e.g., code, data, models), used in the paper, properly credited and are the license and terms of use explicitly mentioned and properly respected?
    \item[] Answer: \answerYes{} 
    \item[] Justification: Localize-MI, SEED, and SEED-IV are cited (\citep{mikulan2020simultaneous, duan2013differential, zheng2015investigating}) with license/access terms reproduced in App.~\ref{app:datasets}; baseline implementations (SaSDim, SADI, RDPI, DDPMEEG, ESTformer, STAD, SRGDiff) are credited to their original papers and used under their original licenses.


\item {\bf New assets}
    \item[] Question: Are new assets introduced in the paper well documented and is the documentation provided alongside the assets?
    \item[] Answer: \answerYes{} 
    \item[] Justification: The released EMAG codebase will be shipped with a README documenting installation, data preparation, training, and evaluation commands.

\item {\bf Crowdsourcing and research with human subjects}
    \item[] Question: For crowdsourcing experiments and research with human subjects, does the paper include the full text of instructions given to participants and screenshots, if applicable, as well as details about compensation (if any)? 
    \item[] Answer: \answerNA{} 
    \item[] Justification: The paper does not involve crowdsourcing or new human-subjects research.

\item {\bf Institutional review board (IRB) approvals or equivalent for research with human subjects}
    \item[] Question: Does the paper describe potential risks incurred by study participants, whether such risks were disclosed to the subjects, and whether Institutional Review Board (IRB) approvals (or an equivalent approval/review based on the requirements of your country or institution) were obtained?
    \item[] Answer: \answerNA{} 
    \item[] Justification: All datasets used were released by the original authors under their own ethics approvals; no additional human-subjects research was conducted.

\item {\bf Declaration of LLM usage}
    \item[] Question: Does the paper describe the usage of LLMs if it is an important, original, or non-standard component of the core methods in this research? Note that if the LLM is used only for writing, editing, or formatting purposes and does \emph{not} impact the core methodology, scientific rigor, or originality of the research, declaration is not required.
    \item[] Answer: \answerYes{} 
    \item[] Justification: LLM coding assistants were used for non-trivial implementation work --- e.g.\ vectorising the 4D Gaussian rendering kernel, writing CUDA-friendly time-chunked variants, and generating test/visualisation code. They were \emph{not} used for research ideation, methodological design, or writing of the scientific narrative; all conceptual and experimental decisions were made by the authors. Drafts produced with LLM assistance were edited by the authors.

\end{enumerate}

\end{document}